\documentclass[twoside,11pt]{article}

\usepackage{jmlr2e}

\usepackage{amsfonts}
\usepackage{amsmath}
\usepackage{amssymb}
\usepackage[title]{appendix}

\usepackage{stmaryrd} 
\usepackage[tight]{subfigure}
\usepackage[pdftex]{hyperref}
\usepackage{subfigmat}
\usepackage{wrapfig}
\usepackage{mathrsfs} 

\usepackage{algorithm}
\usepackage{algorithmic}
\usepackage{color}

\newcommand{\erf}{\operatorname{erf}}

\begin{document}
\title{Parametric Learning and Monte Carlo Optimization}
\author{
\name
David H. Wolpert 
\email 
dhw@email.arc.nasa.gov\\
\addr
MS 269-1, Ames Research Center \\
Moffett Field, CA 94035.
\AND
\name Dev G. Rajnayaran 
\email dgorur@stanford.edu\\
\addr 
Department of Aeronautics and Astronautics,\\
Durand Rm. 158,
496 Lomita Mall,
Stanford, CA 94305.}
\maketitle

\begin{abstract}
This paper uncovers and explores the close relationship between Monte
Carlo Optimization of a parametrized integral (MCO), Parametric
machine-Learning (PL), and `blackbox' or `oracle'-based optimization
(BO). We make four contributions. First, we prove that MCO is
mathematically identical to a broad class of PL problems. This
identity potentially provides a new application domain for all broadly
applicable PL techniques: MCO. Second, we introduce immediate
sampling, a new version of the Probability Collectives (PC) algorithm
for blackbox optimization.  Immediate sampling transforms the original
BO problem into an MCO problem.  Accordingly, by combining these first
two contributions, we can apply all PL techniques to BO. In our third
contribution we validate this way of improving BO by demonstrating
that cross-validation and bagging improve immediate sampling. Finally,
conventional MC and MCO procedures ignore the relationship between the
sample point locations and the associated values of the integrand;
only the values of the integrand at those locations are considered. We
demonstrate that one can exploit the sample location information using
PL techniques, for example by forming a fit of the sample locations to
the associated values of the integrand. This provides an additional
way to apply PL techniques to improve MCO.
\end{abstract}

\section{Introduction}

This paper uncovers and explores some aspects of the close
relationship between Monte Carlo Optimization of a parametrized
integral (MCO), Parametric machine Learning (PL), and `blackbox' or
`oracle-based' optimization (BO). We make four primary
contributions. First, we establish a mathematical identity equating
MCO with PL. This identity potentially provides a new application
domain for all broadly-applicable PL techniques, viz., MCO.

Our second contribution is the introduction of immediate
sampling. This is a new version of the Probability Collectives (PC)
approach to blackbox optimization. PC encompasses Estimation of
Distribution Algorithms (EDAs)\citep{deis97,eda01,lola05} and the
Cross Entropy (CE) method~\citep{rukr04} as special cases. However PC
is broader and more fully motivated. This means it uncovers (and
overcomes) formal shortcomings in those other approaches. 

In the immediate sampling version of PC the original BO problem is
transformed into an MCO problem. In light of our first contribution,
this means we can apply PL to immediate sampling. In this way $all$ PL
techniques --- including cross-validation, bagging, boosting, active
learning, stacking, and others --- can be applied to blackbox
optimization. 

In our third contribution we experimentally explore the power of this
identity between MCO and PL.  In these experiments we demonstrate that
cross-validation and bagging improve the performance of immediate
sampling blackbox optimization. In particular, in these experiments we
show that cross-validation can be used to adaptively set an `annealing
schedule' for blackbox optimization using immediate sampling
{\it{without any extra calls to the oracle}}. In some cases, we show
that this adaptively formed annealing schedule results in better
optimization performance than
\emph{any} exponential annealing schedule.{\footnote{Since
they are special cases of PC, presumably we could similarly apply PL
techniques to improve EDA's or the CE method.}}

Finally, conventional MC and MCO procedures ignore the relationship
between the sample point locations and the associated values of the
integrand. (Only the values of the integrand at the sample locations
are considered by such algorithms.) We end by exploring ways to use PL
techniques to exploit the information in the sample locations, for
instance, by Bayesian fitting of a surface from the sample locations
to the associated values of the integrand. This constitutes yet
another way of applying PL to MCO in general, and therefore to BO in
particular.


\subsection{Background on PL, MCO, Blackbox Optimization, and PC}

We begin by sketching the four disciplines discussed in this paper:

\begin{enumerate}
\item A large number of parametric machine-learning problems share the
following two properties. First, the goal in these problems is to find
a set of parameters, $\theta$, that minimizes an integral of a function that is parametrized by $\theta$. Second, to
help us find that $\theta$ we are are given samples of the
integrand.  These problems reduce to a sample-based search for the $\theta$ that
we predict minimizes the integral. We
will refer to problems of this class as Parametric Learning (PL)
problems.


An example of PL is parametric supervised learning, where we want to find
an optimal predictor or regressor $z_\theta$ that minimizes the associated expected loss, $\int dx\,dy
\; P(x) P(y \mid x) L[y, z_\theta(x)]$, where $x$'s are inputs and
$y$'s are outputs. We do not, however, know $P(x) P(y \mid x)$. 
Instead, we are provided a training set of samples of $P(x)
P(y \mid x)$. The associated PL problem is to use those samples to
estimate the optimal $\theta$.

\item MCO is a technique for solving problems of the form argmin$_\phi
\int dw \; U(w,\phi)$~\citep[see][]{erno98}. MCO starts by replacing that
integral with an importance-sample generated estimate of it. That
estimate is a sum parametrized by $\phi$. In MCO one searches for
the value $\phi$ that minimizes this sum; the result of this search is
one's estimate of the $\phi$ that optimizes the original integral.

\item Blackbox optimization algorithms are ways to minimize functions of
the form $G : X \rightarrow {\mathbb{R}}$ when one does not actually
know the function $G$. Such algorithms work by an iterative process in
which they first select a query $x \in X$, and then an `oracle'
returns to the algorithm a (potentially noise-corrupted) value $G(x)$,
and no other information, in particular, no gradient information. The
difference between one blackbox optimization algorithm and another is
how they select each successive query based on the earlier responses
of the oracle. Examples of blackbox optimization algorithms are
genetic algorithms~\citep{mitc96}, simulated annealing~\citep{kige83},
hill-climbing algorithms, Response-Surface Methods
(RSMs)~\citep{mymo02}, and some forms of Sequential Quadratic
Programming (SQP)~\citep{gimu81,nowr99}, Estimation of Distribution
Algorithms (EDAs)\citep{deis97,eda01,lola05}, tabu search, the Cross Entropy (CE)
method~\citep{rukr04}, and others.

\item PC is a set of techniques that can be used for blackbox
optimization. Broadly speaking, PC works by transforming a search for
the best value of a variable $x$ into a search for the best probability
distribution over the variable, $q(x)$~\citep[see][]{wost06,mawo05,wolp03b,wolp04a,biwo04a,anbi04,lewo04b}. Once
one solves for the optimal $q(x)$, inversion to get the optimal $x$
for the original search problem is stochastic; one simply samples $q$.
As described below, PC has many practical strengths, and is related to 
RSMs, EDAs, and the CE method.
\end{enumerate}

\subsection{Roadmap of This Paper}

We make four primary contributions:
\begin{enumerate}
\item  Sec.~\ref{sec:sup_equals_mco} begins with a detailed review of
MCO and PL. Conventional analysis of Monte Carlo estimation involves a
bias-variance decomposition of the error of the estimator of a
particular integral. We show that for MCO, a full analysis requires
more than simply extending such bias-variance analysis separately to
each of the estimators given by the separate $\phi$'s. Moments
coupling the errors of the estimators for the separate $\phi$'s must
also be taken into account. How should we do that?

To answer this, we note that in a different context, the techniques of
PL take such coupling moments into account, albeit implicitly. This
leads us to explore the relation between MCO and PL. This in turn
leads to our first major contribution, the proof that MCO is identical
to PL. This contribution means that one can apply all PL techniques,
for instance, cross-validation, bagging, boosting, stacking, active
learning and others, to MCO.  Such PL-based MCO (PLMCO) provides a new
way of minimizing potentially high-dimensional parametrized integrals.

Experimentally testing the utility of applying PL to MCO requires
an MCO application domain. Here we choose the domain of blackbox
optimization. To establish how blackbox optimization is an application
domain for MCO requires our second contribution, as follows.

\item We start in Sec.~\ref{sec:pc} by presenting an overview of previous
versions of the blackbox optimization approach of PC. We then make our
second contribution in the following section, where we introduce
immediate sampling, a new version of PC that overcomes some of the
limitations of previous versions.

These first two contributions are combined by the fact that immediate
sampling is a special case of MCO.  The resultant identity between PL
and immediate sampling means that, in principle, any PL technique can be
applied to blackbox optimization. In particular, regularization,
cross-validation, bagging, active learning, boosting, stacking, kernel
machines, and others, can be `cut and paste' to do blackbox optimization.
This use of PL for blackbox optimization constitutes a new application
domain for PL.

In Sec.~\ref{sec:impls} we present some concrete instances of how to
modify immediate sampling to use PLMCO rather than conventional
MCO. It is important to note that when applied (via immediate
sampling) to blackbox optimization, these PL techniques do $not$
require additional calls to the oracle. For example, using
cross-validation to set regularization parameters in immediate
sampling (the equivalent of an annealing schedule in SA) does not
involve running the entire blackbox optimization algorithm with
different regularization schedules.  As another example, using bagging
in immediate sampling does not mean running the optimization algorithm
multiple times based on different subsets of the sample points found
so far.

\item Our third contribution is to experimentally demonstrate  in
Sec.~\ref{sec:exps} that PLMCO substantially outperforms conventional
MCO when used this way for blackbox optimization. We are particularly
interested in blackbox optimization problems where calls to the oracle
are the primary expense. Accordingly, non-oracle, `offline'
computation is considered free. So in our experiments we compare
algorithms based on the values of $G$ found by the algorithms versus
the associated number of calls to the oracle{\footnote{See
~\citet{woma97,Droste2002, woma05,CK03,Igel2004,svw01} for a
discussion of the mathematics relating algorithms under such
performance measures.}}. In particular, we show that bagging and
cross-validation leads to faster blackbox optimization on two
well-known benchmark problems for continuous nonconvex optimization.

It should be emphasized that these experiments are {\it{not}} intended
to investigate whether PLMCO applied to immediate sampling is superior
to other blackbox optimization algorithms. Rather their purpose is to
investigate whether one can indeed leverage the formal connection
between PL and MCO to improve immediate sampling. Accordingly, these
experiments are on toy domains, and we do not compare performance with
other blackbox optimization algorithms. We leave such comparisons to
future papers.

\item In estimating the value of an integral based on random samples
of its integrand, conventional MC and MCO techniques ignore how the
locations of the sample points are related to the associated values of
the integrand. Such techniques concentrate exclusively on those sample
values of the integrand that are returned by the oracle. However, one
can use the sample locations and associated integrand values to form a
supervised learning fit to the integrand. In principle, such a fit can 
then be used to improve the overall estimate of the integral.

In `fit-based' MC and MCO one uses all the data at hand to fit the
integrand and then uses that fit to improve the algorithm.  In
this paper, we concentrate on situations where the data at hand
consist only of sample locations and the associated values of the
integrand, but in other situations the data at hand may also include
information like the gradient of the integrand at the sample
points. In their most general form, fit-based MC and MCO include
techniques to exploit such information.

One natural Bayesian approach to fit-based MC uses Gaussian
processes. Work adopting this approach, for the case where the data
only contain sample locations and associated integrand values, is
reviewed in ~\citet{ragh03}.  In Sec.~\ref{sec:fit_based} we generalize
that work on fit-based MC, e.g., to allow non-Bayesian approaches. In
that section, we also consider fit-based MCO in general, and fit-based immediate
sampling in particular.

\end{enumerate}

One of the ways cross-validation is used in these experiments is to
set a regularization parameter. In immediate sampling, the
regularization parameter plays the same role as the temperature does
in simulated annealing. So intuitively speaking, our results show how
to use cross-validation to set an annealing schedule adaptively for
blackbox optimization, {\it{without extra calls to the oracle}}. We
show in particular that such auto-annealing outperforms the best-fit
exponential annealing schedule.

There are more topics involving the connection between MCO, immediate
sampling and PL, than can be explored in this single paper. One such
topic is how to incorporate constraints on $x$ in immediate
sampling. Another important topic involves a derivation from first principles of
the objective function used in immediate sampling.  These two topics
are briefly discussed in the appendices. Some other topics are
mentioned, albeit even more briefly, in the conclusion.

\subsection{Notation}

As a point of notation, we will use the term `distribution' to
refer either to a probability distribution or a density function, with
the associated Borel field implicitly fixing the meaning. Similarly,
we will write integrals even when we mean sums; the measure of the
integral is implicitly taken to be the one appropriate for the the
argument. We will use $\Theta$ to indicate the Heaviside or indicator
function, which is 1 if its argument is positive, and 0 otherwise. 

We will use $\mathscr{P}$ to mean the set of all distributions over $X$. We
are primarily interested in $X$'s that are too large to permit computations involving all members of
$\mathscr{P}$. Accordingly, we will will work with parametrized
subsets ${\mathscr{Q}} \subset {\mathscr{P}}$. We generically write that (possibly vector-valued)parameter as $\theta$, and write the element of
$\mathscr{Q}$ specified by $\theta$ as $q_\theta$. We use $\mathbb{E}$
to indicate the expectation of a random
variable. Subscripts on $\mathbb{E}$ are sometimes used to indicate
the distribution(s) defining the expectation.

We take any oracle ${\mathscr{G}}$ to be an $x$-indexed set of
independent stochastic processes, and use the symbol $g$ to
indicate the generic output of the oracle in response to any query.
With some abuse of notation, we denote the output of
the oracle for query $x$ as $P(g \mid x, {\mathscr{G}})$. For a noise-free, or single-valued oracle, 
we write $P(g \mid x, {\mathscr{G}}) = \delta(g - G(x))$ for some function $G$ implicitly
specified by $\mathscr{G}$, where $\delta(.)$ is the Dirac delta
function.

When there is both a factual version of a random variable and a
posterior distribution over counter-factual values of that variable,
they must be distinguished. In general this requires extending the
conventional Bayesian formalism~\citep[see][]{wolp97,wolp96c}. Here,
though, it suffices for us to indicate counter-factual values by a
subscript $c$. Say there is a factual oracle $\mathscr{G}$, and we are
provided a data set $D$ formed by sampling $\mathscr{G}$. We use
superscripts to denote different samples in that data set.  Then $D$
in turn induces a posterior over oracles, and we write that posterior
as $P({\mathscr{G}}_c \mid D)$.

\section{MCO and PL}
\label{sec:sup_equals_mco}

In this section we review PL and MCO show that they are mathematically identical.

\subsection{Overview of PL}

A broad class of parametric machine learning problems try to find
\begin{eqnarray}
\mathrm{(P1):}&&{\mbox{argmin}}_\xi \int dx \; P(x) R_x(\xi) .\nonumber
\end{eqnarray}
For subsequent purposes, it will be useful to write $x$ as a subscript and
$\xi$ as an argument of $R$, even though $x$ is the integration
variable and $\xi$ is the parameter being optimized.
To perform this minimization, we have a set of function values $D \equiv \{R_{x^i}(\xi)\}$, where we typically assume that the samples $x^i,i=1,\ldots,N$ were formed by IID sampling of $P(x)$. 

The maximum likelihood approach to this minimization first makes the
approximation
\begin{eqnarray}
\int dx \; P(x) R_x(\xi) &\approx& \frac{1}{N}\sum_i R_{x^i}(\xi), 
 \nonumber\\
&\triangleq& \frac{1}{N}\sum_i R^i(\xi).\label{eq:learn_approx}
\end{eqnarray}
One then solves for the $\xi$ minimizing the sum, and uses this as an
approximation to the solution to P1. In practice, though, this procedure 
is seldom used directly: although the approximation
in Eq.~\ref{eq:learn_approx} is unbiased for any fixed $\xi$, 
min$_\xi \sum_i R^i(\xi)$ is not an unbiased estimate of min$_\xi \int
dx \; P(x) R_x(\xi)$. Therefore, when this approximation is exploited, it is modified to incorporate bias-reduction techniques. 

\subsubsection*{Example: Parametric Supervised Learning:}
Let $X, Y$ be input and output spaces, respectively. Let $L(y^1, y^2) : Y \times Y
\rightarrow {\mathbb{R}}$ be a loss function, and $z_\xi : X
\rightarrow Y$ be a $\xi$-parametrized set of functions. In
parametric machine learning with IID error our goal is to solve

\begin{eqnarray}
{\mbox{argmin}}_\xi && \int dx \; P(x)\int dy \; P(y \mid x) L(y,
z_\xi(x)),\nonumber  \\
&\triangleq& \nonumber \\
{\mbox{argmin}}_\xi && \int dx \; P(x) R_x(\xi).
\label{eq:sup_learn}
\end{eqnarray}
Intuitively, $R_x(\xi)$ is the expected loss at
$x$ for the `fit' $z_\xi(x)$ to the $x$-indexed set of
distributions $P(y \mid x)$.

To perform this minimization we have a {\bf{training set}} of pairs $D
\equiv \{x^i, y^i\},i=1,\ldots,N$, that we assume were formed by
IID sampling of $P(x) P(y \mid x)$.  The maximum likelihood approach
to this minimization first makes the approximation
\begin{eqnarray}
\int dx \; P(x)\int dy \; P(y \mid x) L(y, z_\xi(x)) &\approx&\frac{1}{N}
\sum_i L(y^i, z_\xi(x^i)), \nonumber \\
&\triangleq& \frac{1}{N}\sum_i R^i(\xi).\nonumber
\end{eqnarray}
One then solves for the $\xi$ minimizing the sum, and uses this as an
approximation to the solution to (P1). As discussed above, in practice, this
minimization is rarely used directly, and is usually combined with a bias-reducing technique like cross-validation.

\subsection{Overview of MCO}
\label{sec:mco}

Consider the problem
\begin{eqnarray*}
\mathrm{(P2):}&& {\mbox{argmin}}_{\phi \in \Phi} \int dw \; U(w, \phi).
\label{eq:mco_def}
\end{eqnarray*}
For now, we do not impose constraints on $\phi$, nor restrict $\Phi$. Monte Carlo Optimization~\citep{erno98} is a way to search for the solution of (P2).  In MCO we use importance sampling to rewrite the integral in (P2) as
\begin{eqnarray}
\int dw \; U(w, \phi) &=&  \int dw \; {v}(w) \frac{U(w, \phi)}{{v}(w)}. \nonumber \\
&\triangleq& \int dw  \; {v}(w) r_{{v},U,w}(\phi),
\label{eq:mco_samp}
\end{eqnarray}
for some sampling distribution ${v}$. Following the usual
importance sampling procedure, we IID sample ${v}$ to form a sample set
\{$U(w^i, .) : i = 1, \ldots N$\}, which specifies a set of $N$ sample functions
\begin{eqnarray*}
r^i(\phi) &\triangleq& r_{v,U,w^i}(\phi).
\end{eqnarray*}
It is implicitly assumed that for any $w$, we can evaluate $v(w)$ up
to an overall normalization constant.

In MCO, these $N$ functions are used in combination with any prior
information to estimate the solution to (P2). Conventionally, this is
done by approximating the solution to (P2) with the solution to the
problem
\begin{eqnarray*}
\mathrm{(P3):}&& {\mbox{argmin}}_{\phi} \sum_i r^i(\phi).
\end{eqnarray*}
We define 
\begin{eqnarray}
{\mathscr{L}}_U(\phi) &\triangleq& \int dw \; U(w, \phi),\nonumber  \\
{\hat{\mathscr{L}}_{v,U,\{w^i\}}}(\phi) &\triangleq& \sum_i
r_{v,U,w^i}(\phi) ,\nonumber\\
{\hat{\phi}}_{v,U,\{w^i\}}(\phi) &\triangleq& {\mbox{argmin}} [{\hat{\mathscr{L}}_{v,U,\{w^i\}}}(\phi)].\nonumber
\end{eqnarray}
For notational simplicity, the subscripts will usually be omitted
in these expressions.  We will use the term {\bf{naive MCO}} to
refer to solving (P3) by minimizing ${\hat{\mathscr{L}}}(\phi)$.

\subsection{Statistical Analysis of MCO}
\label{sec:covars}

The statistical analysis of MC estimation of integrals is a relatively
mature field~\citep[see][]{roca04,fish96}. We now show that when such MC estimation is combined with parameter optimization in MCO, the analysis becomes much more involved.

\subsubsection{Review: MC Estimation}
\label{subsubsec:MCEstimation}
First consider MC estimation, with no mention of MCO. We first need to to specify a loss function $L(., .)$ that will couple our mathematics with real-world costs. The
first argument of such an $L$ is the output of the estimation algorithm under
consideration. The second argument is the quantity statistically
sampled by that algorithm. The associated value of $L$ is the cost if
the algorithm produces the output specified in that first argument,
using the quantity specified in the second argument.

As an example, consider importance-sampled MC estimation of an
integral. Using the MCO notation just introduced, we use
${\hat{\mathscr{L}}}(\phi)$ as an estimate of ${\mathscr{L}}(\phi)$
for some fixed $\phi$. The quantity being sampled is the
function $U(., \phi)$, and the output of the algorithm is
${\hat{\mathscr{L}}}(\phi)$. Accordingly, these are the arguments of
the loss function. 

The most popular loss function in statistical analysis of MC integral
estimation is quadratic loss, given below.
\begin{eqnarray*}
L({\hat{\mathscr{L}}}(\phi), U(., \phi)) &\triangleq&
[{\hat{\mathscr{L}}}(\phi) - \int  dw \; U(w, \phi)]^2 .
\end{eqnarray*}
Unless explicitly stated otherwise, we will henceforth use the term
`expected loss' to refer to the average of this loss function over
sample sets.  Since ${\hat{\mathscr{L}}}(\phi)$ is an unbiased
estimate of ${\mathscr{L}}(\phi)$, the expected loss is the sample
variance,
\begin{eqnarray*}
{\mbox{Var}}({\hat{\mathscr{L}}}(\phi)) &=& {\mathbb{E}}([\sum_{j=1}^N \frac{U(w^j, \phi)}{N v(w^j)}]^2) \;\;-\;\; 
[{\mathbb{E}}(\sum_{j=1}^N \frac{U(w^j, \phi)}{N v(w^j)})]^2 \nonumber \\
&=& \frac{1}{N} \{ \int dw \; v(w) [\frac{U(w, \phi)}{v(w)}]^2 \;\;-\;\; [\int dw
\; v(w) \frac{U(w, \phi)}{v(w)}]^2 \} \nonumber \\
&=& \frac{1}{N} \{ \int dw \; v(w) [\frac{U(w, \phi)}{v(w)}]^2 \;\;-\;\;
[{\mathscr{L}}(\phi)]^2 \} .
\end{eqnarray*}
This expansion for the sample variance is quite useful. For example,
one can solve for the $v$ that minimizes this variance (and therefore
minimizes expected loss) as a function of $U(., \phi)$. For nowhere-negative $U$, that optimal $v$ is given by ~\citep[see][]{roca04}
\[v(w) \triangleq
\frac{U(w, \phi)}{\int dw' U(w', \phi)}.\]
Given the formula for the optimal $v$, one can estimate it from a current sample set, and then use the estimated optimal $v$ for future sampling. This is what is done in the VEGAS
Algorithm~\citep{lepa78,lepa80}.  Consideration of the sample variance
has also led to algorithms that partition $X$ and then run importance
sampling on each partition element separately, for instance, stratified
sampling~\citep{fish96}.  MC estimators that do not use strict importance sampling may introduce bias. However, if the variance is sufficiently reduced, 
expected quadratic error is reduced. This can be exploited to
tradeoff bias and variance.

\subsubsection{From MC to MCO}
When we combine MC with parameter optimization in MCO, quantities like ${\mbox{Var}}({\hat{\mathscr{L}}}(\phi))$ for
one particular $\phi$ are not the main objects of interest. Instead,
we are interested in expected loss of our iterated MCO algorithm, which
involves multiple $\phi$'s. So what is the appropriate loss function
for analyzing MCO? From the very definition of (P2), it is clear that
we want $L(\phi,U)$ to be minimized by the $\phi$ that minimizes $\int dw \; U(w, \phi)$. 
The simplest approach to doing this, which will be assumed from now
on, stipulates that
\begin{eqnarray}
L(\phi, U) \;\;=\;\; {\mathscr{L}}(\phi) \;\;=\;\; \int dw \; U(w, \phi) ,
\label{eq:mco_loss}
\end{eqnarray} 
the same integral appearing in (P2).  If we can solve (P2)
exactly, then we will have produced the $\phi$ with minimal value of
this loss function.

Given this choice of loss function, expected loss in naive MCO is
\begin{eqnarray}
{\mathbb{E}}(L \mid U, v) &=& \int dw^1 \ldots dw^N \; \prod_{i=1}^N
v(w^i) {\mathscr{L}} ({\mbox{argmin}}_\phi [{\hat{\mathscr{L}}}_{v,U,\{w^i\}}(\phi)]) \nonumber \\
&=& \int dw^1 \ldots dw^N \; \prod_{i=1}^N
v(w^i) \int dw' \; U(w', {\mbox{argmin}}_\phi [\sum_{j=1}^N
\frac{U(w^i, \phi)}{v(w^i)}]) .
\label{eq:exp_loss_form}
\end{eqnarray}
The optimal $v$ for naive MCO is the one that minimizes
${\mathbb{E}}(L \mid U, v)$. There is no direct relation between this
$v$ and the one that minimizes loss for some single $\phi$. In stark contrast to the MC analysis in Sec.~\ref{subsubsec:MCEstimation},  in addition to the sample variance ${\mbox{Var}}({\hat{\mathscr{L}}}(\phi))$ for any \emph{single} $\phi$,  the expected loss ${\mathbb{E}}(L \mid U, v)$
now also depends on moments coupling the distributions of $\hat{\mathscr{L}}(\phi)$ for \emph{different} $\phi$'s. Loosely speaking, the bias-variance tradeoff in Sec.~\ref{subsubsec:MCEstimation} now becomes a more complicated bias-variance-covariance tradeoff. Now, setting $w$ is more involved, but we can approach it as follows.

Expressing the expected loss slightly differently gives us an
important insight. Note that each sample set $\{w^i\}$ gives rise to
an associated set of estimates for all $\phi \in \Phi$. Call this
(possibly infinite dimensional) vector of estimates $\vec{l}$, each of
whose components is indexed by $\phi$ and is an estimate for that
particular $\phi$. Now, instead of computing expected loss by
averaging over all possible sample sets, we average over all possible
vectors $\vec{l}$. In order to do this, we need to specify the
probability of each vector $\vec{l}$. Define
\begin{eqnarray*}
\pi_{v,U,\Phi}({\vec{l}}) &\triangleq& {\mbox{Pr}}(\{w^i\} :
{\hat{\mathscr{L}}}_{v,U,\{w^i\}}(\phi) = l_\phi \;\; \forall \phi
\in \Phi).
\end{eqnarray*}
So, $\pi_{_{v,U,\Phi}}({\vec{l}})$ is the probability of a
set of sample points \{$w^i$\} such that for each $\phi \in \Phi$, the
associated empirical estimate $\sum_i r_{v,U,w^i}(\phi)$ equals the
corresponding component of $\vec{l}$.  For notational simplicity the 
subscripts of $\pi_{_{v,U,\Phi}}$ will sometimes be omitted.
We can now write Eq.~\ref{eq:exp_loss_form} succinctly as
\begin{eqnarray}
{\mathbb{E}}(L \mid U, v) &=& \int d{\vec{l}} \; \pi({\vec{l}}) \;
{\mathscr{L}}({\mbox{argmin}}_\phi [l_\phi])
\label{eq:exped_loss_pi}
\end{eqnarray}
where `argmin$_\phi [l_\phi]$' means the index $\phi$ of the
smallest component of $\vec{l}$.  The {\bf{risk}} is the difference
between this expected loss and the lowest possible loss. We can write
that risk as
\begin{eqnarray} 
\int d{\vec{l}} \; \pi_{v,U,\Phi}({\vec{l}}) \;
[{\mathscr{L}}({\mbox{argmin}}_\phi [l_\phi]) \;-\;
{\mbox{min}}_\phi [{\mathscr{L}}(\phi)]] .
\label{eq:risk_succinct}
\end{eqnarray}

Our sample set constitutes a set of samples of $\pi_{v,U,\Phi}$
occurring in Eq.~\ref{eq:exped_loss_pi},
This fact can potentially be exploited to dynamically modify $v$ and/or $\Phi$ to
reduce ${\mathbb{E}}(L \mid U, v)$. Indeed, for the simpler case of MC estimation, 
this is essentially the kind of computation done in the VEGAS algorithm mentioned above.
As a practical issue, it may be difficult to update $v$ and/or $\Phi$ using the full formula
Eq.~\ref{eq:exp_loss_form}. Instead, one could approximate that
formula ${\mathbb{E}}(L \mid U, v)$ near a single $\phi$ of interest, e.g., about a current estimate for the optimal $\phi$.

Intuitively though, one would expect that for a fixed set of $\phi$'s,
everything else being equal, it would be advantageous to have small variances of unbiased estimators and large covariances between them. Such considerations
based on the second moments may help one choose quantities like the
sampling distribution $v$.

Such considerations may also help one choose the set of candidate
$\phi$'s, $\Phi$. For example, one way to have large covariances
between the $\phi \in \Phi$ is to have the associated functions over
$w$, \{$U(., \phi) : \phi \in \Phi\}$, all lie close to one another in an appropriate function space
(e.g., according to an $l_\infty$ norm comparing such functions).
However, choosing such a $\Phi$ will tend to mean there is a small
`coverage' of that set of functions,
\{$U(., \phi) : \phi \in \Phi\}$. More precisely, it will tend to prevent the best of those
$\phi$'s from being very good; min$_{\phi \in \Phi}[\int dw \; U(w,
\phi)]$ will not be very low.

This illustrates that, in choosing $\Phi$, there will be a
tradeoff between two quantities: The first quantity is the best
possible performance with any of the $\phi \in \Phi$. The second
quantity is the risk, that is, how close a given MCO algorithm
operating on $\Phi$ is likely to come to that best possible
performance of a member of $\Phi$.  Choosing $\Phi$ to have large
covariances of the (MC estimators based on the) members of $\Phi$, and
in particular to have large covariances with the truly optimal $\phi$,
argmin$_{\phi \in \Phi} {\mathscr{L}}(\phi)$, will tend to result in
low risk. But it will also tend to result in poor best-possible
performance over all $\phi \in \Phi$.

Similarly, one would expect that as the size of $\Phi$ increases,
there would be a greater chance that a sample set for one of the
suboptimal $\phi \in \Phi$ would have low expected loss `by
luck'. This would then mislead one into choosing that suboptimal
$\phi$. So increasing the size of $\Phi$ may increase risk. However
increasing $\Phi$'s size should also improve best possible
performance.  So again, we get a tradeoff.

It may be that such considerations involving the size of $\Phi$ and
the covariances of its members can be encapsulated in a single number,
giving an `effective size' of $\Phi$ (somewhat analogously to the VC
dimension of a set of functions).  Such tradeoffs are specific to the
use of MCO, and do not arise in plain (single-$\phi$) MC. They are in
addition to the usual bias-variance tradeoffs, which still apply to
each of the separate MC estimators. 

An illustrative example of the foregoing is provided in App.~\ref{sec:GaussExample}. A more complete statistical analysis of risk in MCO, including Bayesian considerations, is in Sec.~\ref{sec:fit_based}.

\subsection{PL Equals MCO}

In MCO, we have to extrapolate from the sample set of $w$ values to
perform the integral minimization in Eq.~\ref{eq:mco_samp}. As
discussed above, this can recast as having a set of sample functions
$\phi \rightarrow r^i(\phi)$ that we want to use to estimate the
$\phi$ that achieves that minimization. Similarly, in PL, we have to extrapolate from
a training set of functions $R^i(\xi)$ to minimize the integral $\int
dx \; P(x) R_x(\xi)$.  
Though not usually viewed this way, at the root of this extrapolation problem is the problem of using the
sample functions $\xi \rightarrow R^i(\xi)$ to estimate the minimizer
of Eq.~\ref{eq:sup_learn}.

In addition, the analysis of Sec.~\ref{sec:covars} is closely related
to the PL field of uniform convergence theory. That field can be cast
in the terms of the current discussion as considering a broad class of
$U$'s, $\mathscr{U}$. Its starting point is the establishment of
bounds on how
\begin{eqnarray}
{\mbox{max}}_{v,U\in {\mathscr{U}}} [\int d{\vec{l}} \;
\pi_{v,U,\Phi}({\vec{l}}) \;
\Theta({\mathscr{L}}({\mbox{argmin}}_\phi [l_\phi]) \;-\;
{\mbox{min}}_\phi [{\mathscr{L}}(\phi)] \;-\; \kappa)
\label{eq:vapnik}
\end{eqnarray}
depends{\footnote{As an example, rewrite $w \rightarrow
x, \phi \rightarrow \alpha, v(x) \rightarrow P(x)$. Also choose
${\mathscr{U}}$ to be all functions of the form $U(w, \phi) = U(x,
\alpha)
\triangleq \int dy \; P(y \mid x)(y - F(x, \alpha))^2$ for any function
$F$ and distribution $P(y \mid x)$. Under this substitution,
Eq.~\ref{eq:vapnik} becomes the archetypal uniform convergence theory
problem for regression with quadratic loss.}} on $\kappa$. Of particular interest
is how the function taking $\kappa$ to the associated bound varies
with characteristics of $\mathscr{U}$ and
$\Phi$~\citep[see][]{vapn82,vapn95}. Eq.~\ref{eq:vapnik} should be compared
with Eq.~\ref{eq:risk_succinct}.

All of this suggests that the general MCO problem of 
extrapolation from a sample set of empirical functions to minimize the
integral of Eq.~\ref{eq:mco_samp}, is, in fact, identical to the general
PL problem of extrapolation from a
training set of empirical functions to minimize the integral of
Eq.~\ref{eq:sup_learn}. This is indeed the case. As shown in Table~\ref{table:DictionaryTable}, identify
$\xi \leftrightarrow \phi, x \leftrightarrow w, P(x) \leftrightarrow v(w), R_x(\xi)
\leftrightarrow {r}_{v,w}(\phi), r^i_v(\phi) \leftrightarrow R^i(\xi)$. Then the
integrals in Eq.~\ref{eq:mco_samp} and (P1) become identical. So the
MCO expected loss function in Eq.~\ref{eq:mco_loss} becomes identical
to the PL expected loss. 
Similarly, the sample functions for MCO and PL become identical. 

In particular, in supervised learning, when there is no noise, $P(y
\mid x)$ becomes a single-valued function $y(x)$, and the parametric supervised
learning problem becomes
\begin{eqnarray*}
{\mbox{argmin}}_\xi \int dx \; P(x) [L(y(x), z_\xi(x))]
\end{eqnarray*}
This should be compared to the MCO problem as formulated in
Eq.~\ref{eq:mco_samp}. For the same reasons that direct minimization of Eq.~\ref{eq:learn_approx} is seldom used in PL, we now see that naive MCO will be biased, and should preferably not be used directly.

\begin{table}
\centering
\begin{tabular}{|c|c|}
\hline
\textbf{MCO}&\textbf{PL}\\
\hline
$w$ & $x$\\
$\phi$ & $\xi$\\
$v(w)$ & $P(x)$\\
 $r_{v,w}(\phi)$ &$R_x(\xi)$\\
$r^i_v(\phi)$ & $R^i(\xi)$\\
\hline
\end{tabular}
\caption{Correspondence between PL and MCO.}
\label{table:DictionaryTable}
\end{table}

Note that most sampling theory analysis of PL does not directly
consider the biases and variances of the separate Monte Carlo
estimators for each $\xi$, nor does it directly
consider the moments that couple the distributions of those
estimates. Rather, it considers a different type of bias and variance ---
the bias and variance of an entire algorithm that chooses a
$\xi$ based on associated MC estimates of expected
loss~\citep{wolp97}. In this sense, such PL analysis bypasses the
issues considered in Sec.~\ref{sec:covars}. The bias-variance-covariance approach described in this section might have important implications on PL analysis of learning algorithms, but for the moment, in our exploration of the identity between MCO and PL, we simply use PL-based techniques to reduce the bias or variance of our algorithms.


\section{Review of PC}
\label{sec:pc}

This section cursorily reviews the previously investigated type of
PC. It then briefly discusses the advantages of PC for blackbox
optimization and its relation to other blackbox optimization
techniques.

\subsection{Introduction to PC}
\label{sec:overview}

To introduce PC, consider the general (not necessarily blackbox) optimization
problem
\begin{eqnarray*}
\mathrm{(P4):}&& {\mbox{argmin}}_{x \in X} {\mathbb{E}}(g \mid x, {\mathscr{G}}).
\end{eqnarray*}
For now, we ignore constraints on $x$.  In PC we transform (P4) into the problem
\begin{eqnarray*}
\mathrm{(P5):}&&{\mbox{argmin}}_{q_\theta \in {\mathscr{Q}}}{\mathscr{F}}_{\mathscr{G}}(q_\theta), 
\end{eqnarray*}
for some appropriate function ${\mathscr{F}}_{\mathscr{G}}$. After
solving (P5) we stochastically invert $q_\theta$ to get an $x$ (the
ultimate object of interest), by sampling $q_\theta$. This type of
``randomizing transform'' contrasts with conventional transform
techniques, where inversion is deterministic.

Ideally, ${\mathscr{F}}_{\mathscr{G}}$ should be chosen in a
first-principles manner, based on exactly how $q_\theta$ will be
sampled and how those samples used (see Sec.~\ref{sec:fit_based}). In
practice though, computational considerations might lead one to choose
${\mathscr{F}}_{\mathscr{G}}$ heuristically. Intuitively, such considerations might compel us to choose
${\mathscr{F}}_{\mathscr{G}}$ both so that (P5) is readily easy to
solve, and so that any solution $q_\theta$ to (P5) is concentrated
about the solutions of (P4). Taking the parametrization to be implicit, we often abbreviate
${\mathscr{F}}_{\mathscr{G}}(q_\theta)$ as just
${\mathscr{F}}_{\mathscr{G}}(\theta)$.

In many variants of PC explored to date, ${\mathscr{F}}_{\mathscr{G}}(\theta)$ is an integral
transform\footnote{An instance where this is not the case is with the elite objective
function, described in App. B.} over $X$,
\begin{eqnarray}
{\mathscr{F}}_{\mathscr{G}}(\theta) &\triangleq& \int dx\,dg \; P(g
\mid x, {\mathscr{G}}) F(g, q_\theta(x)). 
\label{eq:mco}\\
&\triangleq& \int dx \; r_{P(g \mid x, {\mathscr{G}})}(x, \theta)
\label{eq:pc_prob}
\end{eqnarray}
As an example of such an integral transform, consider optimization
with a noise-free (single-valued) oracle, $P(g \mid x, {\mathscr{G}}) = \delta(g -
G(x))$, where the transformed objective is the expected value of $(g|x)$ under $x \sim q_{\theta}$. In other words, $\mathscr{F}_{\mathscr{G}} = \mathbb{E}_{q_{\theta}}[G(x)]$. In addition, suppose that $\mathscr{Q} = \mathscr{P}$, that is, $q_{\theta}$ can be \emph{any} distribution. Under fairly weak assumptions, it can be shown that one solution to (P5) is given by the point-wise limit  of Boltzmann distributions,
\begin{equation*}
p^{\star}(x) = \lim_{\beta\rightarrow\infty} p^\beta(x),\;\textrm{where } p^{\beta}(x) \propto
\exp[-\beta G(x)].
\end{equation*}
 
In the case where $\mathscr{Q} \subset \mathscr{P}$, we could choose
$\mathscr{F}_{\mathscr{G}}(\theta)$ to be a measure of the
dissimilarity between such a Boltzmann (or other) `target'
distribution, and a given $q_{\theta}$. For instance, we could use a
Kullback-Leibler (KL) divergence between $q_{\theta}$ and $p^{\beta}$,
which we refer to as ``$pq$'' KL distance:
\begin{eqnarray*}
{\mathscr{F}}_{\mathscr{G}}(\theta) &=& {\mbox{KL}}(p^\beta \;|| \; q_\theta) \nonumber\\
&\triangleq& -\int dx \; p^\beta(x)
{\mbox{ln}}[\frac{q_\theta(x)}{p^\beta(x)}] .
\end{eqnarray*}
In terms of the quantities in Eq.~\ref{eq:mco}, $F(g, q_\theta(x))
\propto e^{-\beta g} {\mbox{ln}}[q_\theta(x)]$, up to an overall additive constant. 
So $r_{P(g \mid x, {\mathscr{G}})}(x,\theta)$ in Eq.~\ref{eq:pc_prob} is the contribution to the KL
distance between $p^\beta$ and $q_\theta$ given by the argument $x$.

To see why this choice of ${\mathscr{F}}_{\mathscr{G}}(\theta)$ is reasonable, first
note that $p^\beta(x)$ is large where $G(x)$ is small. Indeed, as $\beta \rightarrow \infty$, $p^{\beta}$ becomes a
delta function about the $x$(s) minimizing $G(x)$, that is, about the
solution(s) to (P4). Now, suppose that ${\mathscr{Q}}$ is a broad
enough class that it can approximate any sufficiently peaked
distribution. That means that there is a $q_\theta \in {\mathscr{Q}}$
for which ${\mbox{KL}}(p^\beta \;|| \; q_\theta)$ is small for large
$\beta$. In such a situation, the $q_\theta$ solving (P5)
will be highly peaked about the $x$(s) solving (P4). Accordingly, if
we can solve (P5) for large $\beta$, sampling the resultant
$q_\theta$ will result in an $x$ with a low ${\mathbb{E}}(g
\mid x, {\mathscr{G}})$. 


\subsection{Review of Delayed Sampling}

We now present a review of  conventional, {\bf{delayed-sampling}} PC. In this type of PC we exploit characteristics of the parametrization of 
$q_\theta$, and pursue the \emph{algebraic} solution of (P5) as far as
possible, in closed form. At some point, if there remain quantities in this algebraic expression that we cannot evaluate closed-form, we estimate them using Monte Carlo sampling.

As an example, consider a noise-free oracle, and instead of $pq$
Kullback-Leibler distance, choose 
${\mathscr{F}}_{\mathscr{G}}(q_{\theta})$ to be the expected value
returned by the oracle under $q_{\theta}$,
\begin{eqnarray}
{\mathbb{E}}_{q_{\theta}, {\mathscr{G}}}(g) &\triangleq& \int dx\,dg \; g P(g \mid x, {\mathscr{G}})
q_\theta(x) \nonumber \\
&=& \int dx \; G(x) q_\theta(x) 
\label{eq:expected_G}
\end{eqnarray}
where the second equality reflects the fact that we are assuming a
noise-free oracle. To emphasize the fact that we're considering
noise-free oracles{\footnote{Even
though it is noise-free, the oracle $G$ may be a random variable ---
we may not know $G$, and may attempt to predict it
probabilistically from data, in a Bayesian fashion. In such a situation, notation like
`${\mathbb{E}}({\mathscr{G}})$' refers to the expected oracle under our prior
distribution over oracles. So, we use ${\mathbb{E}}_{q_\theta, G}(g)$ rather than
${\mathbb{E}}_{q_\theta}(G)$, even though the latter is the notation 
we used in previous work on PC.}}, we will sometimes write ${\mathbb{E}}_{q_{\theta},
{\mathscr{G}}}(g) = {\mathbb{E}}_{q_{\theta},G}(g)$. While
${\mathbb{E}}_{q_{\theta}, {\mathscr{G}}}(g)$ is a linear function of
$q_\theta$, in general it will not be a linear function of
$\theta$. Accordingly, finding the $\theta$ minimizing
${\mathbb{E}}_{q_{\theta}, {\mathscr{G}}}(g)$ may be a non-trivial
optimization problem. 

Since $q_\theta$ must be a probability distribution, (P5) is
actually a constrained optimization problem, involving $|X|$
inequality constraints \{$q_\theta(x) \ge 0 \; \forall x$\}, and one
equality constraint, $\int dx \; q_\theta(x) = 1$. As discussed
by~\citet{wost06}, such a constrained optimization problem can be
converted into one with no inequality constraints by the use of
barrier function methods. These methods transform the original
optimization problem into a sequence of new optimization problems,
\{(P5)$^i$\}, each of which is easier to solve than the original
problem (P5). Solving those problems in sequence leads to a solution
to the original problem (P5).

Consider applying this method with an entropic barrier for the case where ${\mathscr{F}}_{\mathscr{G}}(q_{\theta}) = {\mathbb{E}}_{q_{\theta}, G}(g)$. Then, it turns out that up to additive
constants, each problem (P5)$^i$ is again of the form of (P5). However
the ${\mathscr{F}}_{\mathscr{G}}(q_{\theta})$ of each problem (P5)$^i$
is the `$qp$' KL distance, KL$(q_\theta \; ||
\; p^{\beta_i})$, where $\beta_i$ is the value of the `barrier parameter'
specifying problem (P5)$^i$.  In other words, up to irrelevant additive
constants, each (P5)$^i$ is the problem of finding the $\theta$ that
minimizes
\begin{eqnarray*}
{\mathscr{F}}_{G,{\beta_i}}(q_\theta) &=& {\mathbb{E}}_{q_\theta, G}(g) - {\beta_i}^{-1} S(q_\theta)
\end{eqnarray*}
where $S(.)$ is conventional Shannon entropy{\footnote{This
$qp$ distance is just the free energy of $q_\theta$ for Hamiltonian
function $G$ and inverse temperature $\beta_i$. This gives a novel derivation of the physics
injunction to minimize the free energy of a system.}}. In this case
the barrier function method directs us to iterate the following
process: Solve for the $q_\theta$ that minimizes KL$(q_\theta \; || \;
p^{\beta_i})$, and then update $\beta_i$. 
At the end of this process we will have a local solution to (P5).

In the case where $X$ is a Cartesian product, we often use distributions parametrized as
a product distribution, $q_\theta = \prod_i q_i(x_i)$. Under this parametrization each problem (P5)$^i$ can be solved by gradient descent, where the gradient components of 
${\mathscr{F}}_{G,{\beta_i}}(q_\theta)$ are given by
\begin{eqnarray*}
\frac{\partial {\mathscr{F}}_{G,{\beta_i}}(q_\theta)} {\partial q_i(x_i)} &=& {\mathbb{E}}_{q_\theta, G}(g \mid x_i) + {\beta_i}^{-1}
{\mbox{ln}} [q_i(x_i)] + \lambda_i
\end{eqnarray*}
where the Lagrange parameters $\lambda_i$ enforce normalization of
each $q_i$. 

There are many better alternatives{\footnote{One of these alternatives can be
cast as a corrected version of the replicator dynamics of evolutionary
game theory~\citep{wolp04c}. This may have interesting implications for GAs, which presume evolutionary processes.}} to simple gradient
descent for minimizing each ${\mathscr{F}}_{G,{\beta_i}}(q_\theta)$,
involving Newton's method, block relaxation, and related
techniques~\citep{wost06}. In all such schemes investigated to date, 
we need to repeatedly evaluate terms like
${\mathbb{E}}_{q_\theta, G}(g \mid x_i)$.  Sometimes that evaluation
can be done closed form~\citep{mawo05,mawo04a}.  In blackbox
optimization though, this is not possible.

Typically, when we cannot evaluate the terms ${\mathbb{E}}_{q_\theta,G}
(g \mid x_i)$ in closed-form we use MC to estimate them. Since we have a 
product distribution, we can generate samples of the joint distribution $q_{\theta}(x)$
by sampling each of the marginals $q_i(x_i)$ separately. One can use those sample
$x$'s as queries to the oracle. Then, by appropriately averaging the
oracle's responses to those queries, one can estimate each term ${\mathbb{E}}_{q_\theta,G} (g \mid x_i)$~\citep{wobi04a}. The product factorization implies that our iterative procedure can be performed in a completely decentralized manner, with a separate program controlling each component $x_i$, and communicating \emph{only} with the oracle\footnote{Each such program may be thought of as an `agent' who updates his probability distribution, and this `collective' of agents performs optimization in a decentralized manner. This led to the name `Probability Collective'.}.

In this scheme, once $q_\theta$ is modified, samples of the oracle
that were generated from preceding $q_\theta$'s can no longer be directly used
to estimate the terms ${\mathbb{E}}_{q_\theta,G} (g \mid x_i)$. However, there are
several `data-aging' heuristics one can employ to reuse such old
data by down-weighting it. 

In all these schemes, while we ultimately use Monte Carlo in the PC,
it is delayed as long as possible in the course of solving (P5). This
is the basis for calling this variant of PC `delayed sampling'.

\subsection{Advantages of PC}
\label{subsec:AdvantagesOfPC}

The PC transformation can substantially alter the optimization landscape. 
For a noise-free oracle $G(x)$, (P4) reduces to the problem of finding
the $x$ that minimizes $G(x)$. In contrast, (P5) is the problem of
finding the $\theta$ that minimizes
${\mathscr{F}}_{\mathscr{G}}(\theta)$. The characteristics of the
problems of minimizing $G(x)$ and minimizing
${\mathscr{F}}_{\mathscr{G}}(\theta)$ can be vastly different.  For
example, suppose $q_{\theta}$ is log-concave in its parameters,
and ${\mathscr{F}}_{\mathscr{G}}$ is $pq$
KL distance. In this case, regardless of the function
$G(x)$, ${\mathscr{F}}_{\mathscr{G}}(\theta)$ is a convex function
of $\theta$, over ${\mathscr{Q}}$. So the PC
transformation converts a problem with potentially many local
minima into a problem with none. See~\citet{wost06} for a discussion
of the geometry of the surface ${\mathscr{F}}_{\mathscr{G}}(\theta) :
\theta \rightarrow {\mathbb{R}}$.

Since it works directly on distributions, PC can handle arbitrary data types.
$X$ can be categorical, real-valued, integer-valued, or a mixture of all of these,  but
in each case, the distribution over $X$ is parametrized by a vector of real numbers. This means
that all such problems `look the same' to much of the mathematics of
PC. Moreover, PC can exploit extremely well-understood techniques (like gradient descent) for optimization of continuous functions of real-valued vectors, and apply them to problems in these arbitrary spaces.

Optimizing over distributions can give sensitivity information:
The distribution $q_\theta$ produced in PC will typically be tightly peaked along
certain directions, while being relatively flat along other directions. This
tells us the relative importance of getting the value of $x$ along those different
directions precisely correct.

We can set the initial distribution for PC to be 
a sum of broad peaks, each centered on a solution 
produced by some other optimization algorithm. Then, as that initial
distribution gets updated in the PC algorithm, the set of solutions
provided by those other optimization algorithms are in essence
combined, to produce a solution that should be superior to any of them
individually. 

Yet another advantage to optimizing a distribution is
that a distribution can easily provide multiple solutions to the
optimization problem, potentially far apart in $X$. Those solutions
can then be compared by the analyst in a qualitative fashion.

As discussed later, there are other advantages that accrue
specifically if one uses the immediate-sampling variant of PC. These
include the ability to reuse all old data, the ability to exploit
prior knowledge concerning the oracle, and the ability to leverage
PL techniques. See~\citet{wost06} for a discussion of other advantages of PC, 
in particular in the context of distributed control.

\subsection{Relation to Other Work}

PC is related to several other optimization techniques.  Consider, for
instance, Response Surface Models (RSM)s~\cite{josc98}. In these
techniques, one uses Design of Experiments (DOE) to evaluate the
objective function at a set of points. Then, a low-order parametric
function, often a quadratic, is fitted to these function
values. Optimization of this `response surface' or `surrogate model'
is considered trivial compared to the original optimization. The
result of this surrogate optimization is then used to get more samples
of the true objective at a different set of points. This procedure is
then iterated using some heuristics, often in conjunction with
trust-region methods to ensure validity of the low-order
approximation. We note the similarities with PC in
Table~\ref{table:RSMandPC}.
\begin{table}
\begin{tabular}{|l|l|}
\hline
\multicolumn{1}{|c|}{RSM}&\multicolumn{1}{|c|}{PC}\\
\hline
Fit parametric function to & Fit parametric distribution to \\
objective function values & target distribution\\
\hline
Heuristics to grow trust region & Cross-validation for regularization\\
\hline
DOE for sample points & Random sampling for sample points\\
\hline
Axis alignment of stencil matters & Parametrization can address axis alignment\\ 
\hline
Surrogate minimization not always easy & Implicit, probabilistic `minimization' of surrogate\\
\hline
\end{tabular}
\caption{Relation to RSM.}
\label{table:RSMandPC}
\end{table}

As another example, some variants of PC exploit MC techniques as
discussed above, and thus stochastically generate populations of
samples.  In their use of random populations these variants of PC are
similar to simulated annealing~\citep{kige83}, and even more so to
techniques like EDA's~\cite{eda01,deis97,lola05} and the CE
method~\citep{rukr04}. However, these other approaches do not
explicitly pursue the optimization of the underlying distribution
$q_\theta$, as in (P5). Accordingly, those approaches cannot exploit
situations in which (P5) can be solved without using a stochastically
generated population~\citep{mawo05,mawo04a}.
See~\citet{mawo05} for a more extensive discussion of the relation of
PC to other techniques.

\section{Immediate sampling}

This section introduces a new PC technique called immediate sampling, and cursorily compares it to delayed sampling. As we have just described, in delayed sampling, we use algebra for as long as possible in our solution of (P5). When closed-form expressions can no longer be evaluated, 
we resort to MC techniques. In {\bf{immediate sampling}}, we form an MC sample
immediately, rather than delaying it as long as possible. That sample gives us an approximation to our objective, $\mathscr{F}_{\mathscr{G}}(\theta)$ for all $\theta \in
{\mathscr{Q}}$. We then search for the $\theta$ that optimizes that
sample-based approximate objective.

\subsection{The General Immediate-sampling Algorithm}

We begin with an illustrative example. Consider an integral transform
${\mathscr{F}}_{\mathscr{G}}(q_\theta)$, and use importance sampling
to rewrite it as
\begin{eqnarray}
\int dx  \; h^1(x) \frac{\int dg \; P(g \mid x, {\mathscr{G}}) 
    F(g, q_\theta(x))}{h^1(x)} &=& \int dx dg \; h^1(x)  P(g \mid x,
   {\mathscr{G}}) \frac{F(g, q_\theta(x))}{h^1(x)}, 
\label{eq:imm_sampl} \\
&\triangleq& \int dx  \; h^1(x) r_{P(g \mid x, {\mathscr{G}}), h^1}(\theta).
\label{eq:imm_sampl_2} 
\end{eqnarray}
where we call $h^1(x)$ the {\bf{sampling distribution}}. 
Note that the $r$ in Eq.~\ref{eq:imm_sampl_2} differs from the one defined in
Eq.~\ref{eq:pc_prob}, as indicated by the extra subscript. This new
$r$ is used in the next section.

We form a {\bf{sample set}} of $N$ pairs $D^1 \equiv \{x^i, g^i\}$ by IID
sampling the distribution $h^1(x)P(g \mid x, {\mathscr{G}})$ 
in the integrand of Eq.~\ref{eq:imm_sampl}. That sampling is the
`immediate' Monte Carlo process. $D^1$ is equivalent to a set of
$N$ {\bf{sample functions}}
\begin{eqnarray*}
r^i_{h^1}(x^i, \theta) \triangleq \frac{F(g^i, q_\theta(x^i))}{h^1(x^i)} \; :
\; i = 1, \ldots N.
\end{eqnarray*}

In the simplest version of immediate sampling, 
we would now use the functions $r^i_{h^1}(x^i,
\theta)$, together with our prior knowledge (if any), to estimate the $\theta$
that minimizes ${\mathscr{F}}_{\mathscr{G}}(q_\theta)$.  As an
example, not using any prior knowledge, we could
estimate ${\mathscr{F}}_{\mathscr{G}}(q_\theta)$ for any $\theta$ as
\begin{eqnarray}
\int dx  \; h^1(x) r_{P(g \mid x, {\mathscr{G}}),h^1}(\theta)
 &\approx& \frac{\sum_{i} r^i_{h^1}(x^i, \theta)} {N} .
\label{eq:max_like}
\end{eqnarray}
This estimate is both an unbiased estimate of
${\mathscr{F}}_{\mathscr{G}}(q_\theta)$ and the maximum likelihood
estimate of ${\mathscr{F}}_{\mathscr{G}}(q_\theta)$. Moreover, it has
these attributes for all $\theta$. 
(This is the advantage of estimating ${\mathscr{F}}_{\mathscr{G}}(q_\theta)$ using importance
sampling with a proposal distribution $h$ that doesn't vary with
$\theta$.  )
Accordingly, to estimate the $\theta$ that minimizes
${\mathscr{F}}_{\mathscr{G}}(q_\theta)$ we could simply search for the
$\theta$ that minimizes $\sum_{i} r^i_{h^1}(x^i,
\theta)$.{\footnote{Since $q_{\theta}$ is normalized, so is  
$\int dx \;F(G(x), q_\theta(x)) \; = \; \frac{\int dx \;F(G(x),
q_\theta(x))}{\int dx \; q_\theta(x)}$.  In Eq.~\ref{eq:max_like} we
fix the denominator integral to 1. In practice though, it may make
sense to replace both of the integrals in this ratio with importance
sample estimates of them. That means dividing the sum in
Eq.~\ref{eq:max_like_final} by $\sum_i q(x^i)/h(x^i)$ and then finding
the $\theta$ that optimizes that ratio of sums, rather than the
$\theta$ that just optimizes the numerator
term~\citep[see][]{roca04}. For example, this can be helpful when one
uses cross-validation to set $\beta$, as described below.}}


Once again, even though the average in Eq.~\ref{eq:max_like} is an
unbiased estimate of ${\mathscr{F}}_{\mathscr{G}}(q_\theta)$ for any
fixed $\theta$, its minimizer is not an unbiased estimate of
min$_\theta {\mathscr{F}}_{\mathscr{G}}(q_\theta)$. This is because
searching for the minimizing $\theta$ introduces bias. Therefore, one
should use some other technique than directly minimizing the righthand
side of Eq.~\ref{eq:max_like} to estimate argmin$_\theta
{\mathscr{F}}_{\mathscr{G}}(q_\theta)$.

\subsection{Immediate Sampling with Multiple Sample Sets}

In general, we will not end the algorithm after forming a single
sample set $D^1$. Instead we will use a map $\eta$ that takes 
$D^1$ 
to a new sampling distribution, $h^2$.  We then
generate new $(x, g)$ pairs using $h^2$, giving us a new sample set
$D^2$.  We then iterate this process until we decide to end the
algorithm, at which point we use all our samples sets together to
estimate argmin$_\theta {\mathscr{F}}_{\mathscr{G}}(q_\theta)$.

To illustrate this we first present an example of a
$\theta$-estimation procedure we could run at the end of the immediate
sampling algorithm. This example is just the extension of the maximum
likelihood $\theta$-estimation procedure introduced above to
accommodate multiple sample sets. Let $N$ be the total number of samples, drawn from $M$ sample sets, with $N_j$ samples in the $j$'th sample set. Let
$h^j$ be sample distribution for the $j$'th sample set, and $r^{i,j}$
the sample function value for the $i$'th element of the $j$'th sample set.  Also
define $D^j
\equiv\{x^{i,j},g^{i,j} : i = 1,
\ldots N_j\}$. Then $\sum_{i=1}^{N_j} r^{i,j}_{h^j}(x^{i,j}, \theta) /
N_j$ is an unbiased estimate of
${\mathscr{F}}_{\mathscr{G}}(q_\theta)$ for any sample set
$j$. Accordingly, any weighted average of these estimates is an unbiased
estimate of ${\mathscr{F}}_{\mathscr{G}}(q_\theta)$:
\begin{eqnarray}
{\mathscr{F}}_{\mathscr{G}}(q_\theta)
 &\approx& \displaystyle \displaystyle \sum_{j=1}^{M}w_j \sum_{i=1}^{N_j} \frac{ r^{i,j}_{h^j}(x^{i,j}, \theta)} {N_j}.
\label{eq:max_like_final}
\end{eqnarray}
Modulo unbiasedness concerns, we could then use the minimizer of Eq.\ref{eq:max_like_final} as our estimate of argmin$_\theta {\mathscr{F}}_{\mathscr{G}}(q_\theta)$.

Say we have fixed on some such $\theta$-estimation procedure to run at
the end of the algorithm. The final step of each iteration of immediate sampling
is to run $\eta$, the map taking the samples generated so far to a new $h$. Ideally, we
want to use the $\eta$ that, when repeatedly run during the algorithm,
maximizes the expected accuracy of the final $\theta$-estimation.
However even for a simple $\theta$-estimation procedure, determining
this optimal $\eta$ can be quite difficult. As discussed later, it is
identical to the active learning problem in machine learning.

In this paper we adopt a two-step heuristic for setting $\eta$. In
the first step, at the end of each iteration, we estimate the optimal $q_\theta$ based on all the sample sets generated so far, using Eq.~\ref{eq:max_like_final}.
In the second step, we complete $\eta$ by setting the new $h$ to the 
current estimate of the optimal $q_\theta$. At that point, the new $h$ is used to generate a new sample
set, and the process repeats.

\subsection{Immediate Sampling with MCMC}

For certain types of ${\mathscr{F}}_{\mathscr{G}}$, it is
possible to form samples using other sampling methods like Markov Chain Monte Carlo (MCMC)~\citep[see][]{mack03,besm00,berg85}. For example, if
${\mathscr{F}}_{\mathscr{G}}(\theta)$ is $pq$ distance from the
Boltzmann distribution $p^{\beta}$ to $q_\theta$, then we can use MCMC to form
a sample set of $p^{\beta}$ (not of $q_\theta$). We can then use that sample
set to form an unbiased estimate of
${\mathscr{F}}_{\mathscr{G}}(\theta)$ for any $\theta$. But if $\beta$ were to change, these old samples cannot be used directly. One would have to resort to additional techniques like rejection sampling in order to reuse these samples. The advantage of using importance sampling is that all previous samples can be reused by the simple expedient of modifying their likelihood ratios. Therefore, in this paper, we only consider sampling distributions $h$ that can be sampled directly, without any need for techniques like MCMC.


\subsection{Advantages of Immediate-Sampling PC}

In contrast to delayed sampling, immediate sampling usually presents 
no difficulty with reusing old data, as shown above; all $(x^{i,j},
g^{i,j})$ pairs can be used directly. Note that
we can also reuse data that was generated when $F$ was different,
for instance, data generated under a differerent $\beta^i$ during a KL distance 
minimization procedure. As
long as we store $h^j(x^{i,j})$ in addition to $g^{i,j}$ and $x^{i,j}$ for every sample,
we can always evaluate $r^{i,j}_{h^j}(x^{i,j}, \theta)$ for any $F$. 

Indeed, we can even comment on optimal ways of reusing this old data.
Since each $r^{i,j}_{h^j}(x^{i,j}, \theta)$ is an unbiased
estimate of the integral ${\mathscr{F}}_{\mathscr{G}}(\theta)$, any
weighted average of the $r^{i,j}_{h^j}(x^{i,j},
\theta)$'s is also an unbiased estimate. This can be exploited in the
$\theta$-estimation procedure.
For instance, consider the estimator of Eq.~\ref{eq:max_like_final}. If 
we have good estimates of the variances of the individual
$r^{i,j}_{h^j}(x^{i,j}, \theta)$, we can weight the terms
$r^{i,j}_{h^j}(x^{i,j}, \theta)$ to minimize the variance of the
associated weighted average estimator. Those weights are proportional
to the inverses of the variances~\citep[see][]{mawo05,lepa78,lepa80}. 
As discussed in Sec.~\ref{sec:covars}, the
accuracy of the associated MCO algorithm could be expected to improve
under such weighting. 

We can also shed light on how to go about gathering new data.
As in the VEGAS Algorithm~\citep{lepa78,lepa80}, one could incorporate 
bias-variance considerations into the
operator $\eta$ that sets the next sampling distribution. To give an
example, let $\Xi$ be the range of $\eta$, and fix $\theta$.  Given
$\Xi$ and $\theta$, one can ask what proposal distribution $h \in
\Xi$ would minimize the sample variance of the estimator in
Eq.~\ref{eq:max_like}. Intuitively, this is akin to asking how best
to do active learning. In general, the answer to this question, the
optimal sampling distribution $h(x)$, will be set by the function $r_{P(g \mid x,
{\mathscr{G}}),h}(\theta)$, viewed as mapping $X \rightarrow
{\mathbb{R}}$.  Accordingly, for any fixed $\theta$, one can use the
MC samples generated so far to estimate the $x$-dependence of $r_{P(g
\mid x, {\mathscr{G}}),h}(\theta)$, and thereby estimate the
optimal $h \in \Xi$. One then uses that estimate as the next sampling distribution $h$.

Another advantage of immediate sampling over delayed sampling is that
the analysis in delayed sampling relies crucially on the parametrization of the
$q$'s; some such parametrizations will permit the
closed-form calculations of delayed sampling, and others will not. In
immediate sampling, this problem disappears.

\subsection{Implications of the Identity Between MCO and PL}
\label{sec:impls}

For the case where ${\mathscr{F}}_{{\mathscr{G}}}(\theta)$ is an
integral transform like Eq.~\ref{eq:mco}, the PC optimization problem
(P5) becomes a special case of minimizing a parametrized integral, the problem (P2). 
Formally, the equivalence is made by equating $x$ with the parameter $\phi$, $g$ with $w$, and $g
\times P(g \mid x, {\mathscr{G}})$ with $U(w, \phi)$. In particular,
immediate sampling is a special case of MCO.
This identity means that we can exploit the extremely well-researched
field of PL to improve many aspects of immediate sampling. In particular:
\begin{itemize}

\item PL techniques like boosting~\citep{scsi99} and bagging~\citep{brei94} can be used 
in (re)using old samples before forming new ones.

\item Variants of active learning{\footnote{Active learning in the precise machine learning sense
uses current data to decide on a new query $x$ to feed to the
oracle. We use the term more loosely here, to refer to any scheme for
using current data to dynamically modify a process for generating for
future queries.}}
can be used to set and update
$h$. Some aspect of this are discussed in Sec.~\ref{sec:fit_based}
below.
\item Cross-validation is directly applicable in many ways: In our context, the curse of dimensionality arises if ${\mathscr{Q}}$ is very large. We can address this the conventional
PL way, by adding a regularization function of $q_\theta$ to the objective function.  
The parameters controlling this regularization can be updated dynamically, as new data is generated, using cross-validation

To use cross-validation this way, one forms multiple partitions of the
current data. For each such partition, one calculates the optimal $q_\theta$ for the training
subset of that partition. One then examines error on the validation
subset of that partition. More precisely, one calculates the \emph{unregularized} objective value on
the held-out data. 

\item More generally, we can use cross-validation to dynamically update \emph{any}
parameters of the immediate sampling algorithm. For example, we can
update the `temperature' parameter $\beta$ of the Boltzmann distribution, 
arising in both $qp$ and $pq$ KL distance, this way.

Note that doing this does not involve making more calls to the
oracle. 

\item We can also use cross-validation to choose the best model class (parametrization)
for $q_{\theta}$, among several candidates.

\item As an alternative to all these uses of cross-validation, one can
use stacking to dynamically combine differrent
temperatures, different parametrized density functions, and so on.

\item One may also be able to apply kernel methods to do the density
estimation~\citep[see][]{macr05}.

\end{itemize}

\section{Experiments}
\label{sec:exps}

In this section, we demonstrate the application of PL and immediate-sampling PC techniques to the unconstrained optimization of continuous functions, both deterministic and nondeterministic. We first describe our choice of $\mathscr{F}_{\mathscr{G}}$, in this case $pq$ KL distance. Next, as an illustrative example, we apply immediate sampling to the simplest of optimization problems, where the objective is a 2-D quadratic. Subsequently, we apply it to deterministic and stochastic versions of two well-known unconstrained optimization benchmarks, the Rosenbrock function and the Woods function. 

We highlight the use of PL techniques to enhance optimizer performance on these benchmark problems. In particular, we show that cross-validation for regularization yields a performance improvement of an order of magnitude. We then show that cross-validation for model-selection results in improved performance, especially in the early stages of the algorithm. We also show that bagging can yield significant improvements in performance. 

\subsection{Minimizing $pq$ KL Distance}
Recall that the integral form of $pq$ KL distance is
\begin{equation*}
\textrm{KL}(p\|q) = \int dx\,p(x) \ln\left(\frac{p(x)}{q(x)}\right).
\end{equation*}
It is easy to show that when there are no restrictions on $q$ being a parametrized density, $pq$ KL distance is minimized if $p=q$. However, owing to sampling considerations, we usually choose $q$ to be some parametric distribution $q_{\theta}$. In this case, we want to find the parameter vector $\theta$ that minimizes KL$(p\|q_{\theta})$. Since the target distribution $p$ is derived purely from $\mathscr{G}$ and is independent of $q_{\theta}$, minimizing $pq$ KL distance is equivalent to the following cross-entropy minimization problem.
\begin{equation}
\label{eq:CEMinimization}
\begin{array}{ll}
\textrm{minimize}&-\int dx\, p(x)\ln\left(q(x)\right),\\
\textrm{subject to}&\int dx\,q(x) = 1,\\
   & q(x) \geq 0 \;\;\;\forall x.
\end{array}
\end{equation}

\subsubsection{Gaussian Densities}
If $q$ is log-concave in its parameters $\theta$, the minimization problem (\ref{eq:CEMinimization}) is a convex optimization problem. In particular, consider the case where $X = \mathbb{R}^n$, and $q_{\theta}$ is a multivariate Gaussian density, with mean $\mu$ and covariance $\Sigma$, parametrized as follows,
\begin{equation*}
q_{\mu,\Sigma}(x) = \frac{1}{(2 \pi)^{n/2} |\Sigma|^{1/2}}\exp\left(-\frac{(x-\mu)^T\Sigma^{-1}(x-\mu)}{2}\right),
\end{equation*}
then the optimal parameters are given by matching first and second moments of $p$ and $q_{\theta}$. 
\begin{equation*}\label{eq:MomentMatching}
\begin{array}{rcl}
\mu^{\star}&=&\int dx\, x\, p(x),\\
\Sigma^{\star}&=&\int dx\, (x-\mu^{\star})(x-\mu^{\star})^T p(x).
\end{array}
\end{equation*}
It is easy to generalize this to the case where $X \subset \mathbb{R}^n$, by making a suitable modification to the definition of $p$. This is described in Sec.~\ref{subsubsec:quadraticG}.

\subsubsection{Immediate Sampling with a Single Gaussian}
Using importance sampling, we can convert the cross-entropy integral in Eq.~\ref{eq:CEMinimization} to a sum over data points, as follows.
\begin{equation*}
\frac{1}{N}\sum_{D} \frac{p(x^i)}{h(x^i)}\ln\left(q_{\theta}(x^i)\right),
\end{equation*}
where $D$ is the data set $\{(x^i,g^i)\},i=1,\ldots,N$. This sets up the minimization problem for immediate sampling for $pq$ KL distance.
\begin{equation}\label{eq:ImmediateSamplingCostFunction}
\textrm{minimize }-\sum_{D} \frac{p(x^i)}{h(x^i)}\ln\left(q_{\theta}(x^i)\right).
\end{equation}
Denote the likelihood ratios by $s^i = p(x^i)/h(x^i)$. Differentiating Eq.~\ref{eq:ImmediateSamplingCostFunction} with respect to the parameters $\mu$ and $\Sigma^{-1}$ and setting them to zero yields\footnote{\textbf{Remarks:}
\begin{enumerate}
\item
As expected, these formul{\ae}, in the infinite-data limit, are identical to the moment-matching results for the full-blown integral case.
\item
The formul{\ae} resemble those for MAP density estimation, often used in supervised learning to find the MAP parameters of a distribution from a set of samples. The difference in this case is that each sample point is weighted by the likelihood ratio $s^i$, and is equivalent to `converting' samples from $h$ to samples from $p$.
\end{enumerate}}

\begin{eqnarray*}
\mu^{\star} &=& \displaystyle\frac{\sum_{D} \displaystyle s^i x^i}{\sum_{D} s^i}\\
\Sigma^{\star} &=& \frac{\sum_{D} s^i (x^i - \mu^{\star})(x^i - \mu^{\star})^T}{\sum_{D} s^i}
\end{eqnarray*}

\subsubsection{Mixture Models}
The single Gaussian is a fairly restrictive class of models. Mixture models can significantly improve flexibility, but at the cost of convexity of the KL distance minimization problem. However, a plethora of techniques from supervized learning, in particular the Expectation Maximization (EM) algorithm, can be applied with minor modifications. 

Suppose $q_{\theta}$ is a mixture of $M$ Gaussians, that is, $\theta = (\mu, \Sigma, \phi)$ where $\phi$ is the mixing p.m.f, we can view the problem as one where a hidden variable $z$ decides which mixture component each sample is drawn from. We then have the optimization problem
\begin{equation*}
\textrm{minimize } -\sum_{D} \displaystyle\frac{p(x^i)}{h(x^i)}\ln\left(q_{\theta}(x^i, z^i)\right).
\end{equation*}
Following the standard EM procedure, we multiply and divide the quantity inside the logarithm by some $Q_i(z^i)$, where $Q_i$ is a distribution over the possible values of $z^i$. As before, let $s^i$ be the likelihood ratio of the $i$'th sample.
\begin{equation*}
\textrm{minimize }-\sum_{D} \displaystyle s^i\ln\left(\sum_{z^i}
      Q_i(z^i)\frac{q_{\theta}(x^i, z^i)}{Q_i(z^i)}\right)
\end{equation*}
Then using Jensen's inequality, we can take $Q_i$ outside the logarithm to get a lower bound. To make this lower bound tight, choose $Q_i(z^i)$ to be the constant $p(z^i|x^i)$. Finally, differentiating with respect to $\mu_j, \Sigma^{-1}_j$ and $\phi_j$ gives us the EM-like algorithm:
\begin{equation*}
\begin{array}{lrcl}
\textrm{E-step: For each i, set} & Q_i(z^i) &=& p(z^i|x^i),\\
\hspace{0.5in}\textrm{that is,}      & w^i_j &=& q_{\mu, \Sigma, \phi}(z^i = j|x^i) ,\;\;\;j=1,\ldots,M.\\
\textrm{M-step: Set} & \mu_j &=& \displaystyle\frac{\sum_{D} w^i_j \displaystyle s^i\,x^i}{\sum_{D} w^i_j \displaystyle s^i},\\
   &\Sigma_j &=& \displaystyle\frac{\sum_{D} w^i_j \displaystyle s^i\,(x^i - \mu_j)(x^i - \mu_j)^T}{\sum_{D} w^i_j \displaystyle s^i},\\
&\phi_j &=& \displaystyle\frac{\sum_{D} w^i_j \displaystyle s^i}{\sum_{D} \displaystyle s^i},\\
\end{array}
\end{equation*}
Since this is a nonconvex problem, one typically runs the algorithm multiple times with random initializations of the parameters.

\subsection{Implementation Details}
In this section we describe the implementation details of an iterative immediate-sampling PC algorithm that uses the Gaussian mixture models described in the previous section to minimize $pq$ KL distance to a Boltzmann target parametrized by $\beta$. We also describe the modification of a variety of techniques from parmetric learning that significantly improve performance of this algorithm. An overview of the procedure is presented in Algorithm~\ref{algo:GaussMixOverview}.
\begin{algorithm}[h]
\caption{Overview of $pq$ minimization using Gaussian mixtures}
\label{algo:GaussMixOverview}
\begin{algorithmic}[1]
\STATE{Draw uniform random samples on $ X $}
\STATE{Initialize regularization parameter $\beta$}
\STATE{Compute $G(x)$ values for those samples}
\REPEAT
\STATE{Find a mixture distribution $q_{\theta}$ to minimize sampled $pq$ KL distance}
\STATE{Sample from $q_{\theta}$}
\STATE{Compute $G(x)$ for those samples}
\STATE{Update $\beta$}
\UNTIL{Termination}
\STATE{Sample final $q_{\theta}$ to get solution(s).}
\end{algorithmic}
\end{algorithm}

\subsubsection{Example: Quadratic $G(x)$}
\label{subsubsec:quadraticG}
\begin{wrapfigure}{R}{0.3\linewidth}
\centering\includegraphics[width=2in]{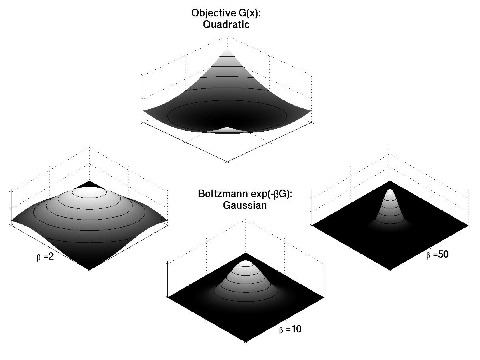}
\caption{\protect \centering Quadratic $G(x)$ and associated Gaussian targets}
\label{fig:QuadraticSurface}
\end{wrapfigure}
Consider the 2-D box $ X  = \{x\in \mathbb{R}^2 \;\;\;|\;\;\; \|x\|_{\infty} < 1\}$. Consider a simple quadratic on $ X $, 
\begin{equation*}
G_Q(x) = x_1^2 + x_2^2 + x_1 x_2, \;\;\; x\in X .
\end{equation*}
The surface and contours of this simple quadratic on $ X $ are shown in Fig.~\ref{fig:QuadraticSurface}. Also shown are the corresponding Boltzmann target distributions $p^{\beta}$ on $ X $, for $\beta = 2, 10$ and 50. As can be seen, as $\beta$ increases, $p^{\beta}$ places increasing probability mass near the optimum of $G(x)$, leading to progressively lower $\mathbb{E}_{p^{\beta}} G(x)$. Also note that since $G(x)$ is a quadratic, $p^{\beta}(x) \propto \exp(-\beta G(x))$ is a Gaussian, restricted to $ X $ and renormalized. We now `fit' a Gaussian density $q_{\theta}$ to the Boltzmann $p^{\beta}$ by minimizing KL($p^{\beta}\|q_{\theta}$), for a sequence of increasing values of $\beta$. Note that $q_{\theta}$ is a distribution over $\mathbb{R}^2$, and $G_Q$ is not defined everywhere in $\mathbb{R}^2$. Therefore, we extend the definition of $G_Q$ to all of $\mathbb{R}^2$ as follows.
\begin{equation*}
G_Q(x) = \left\{
\begin{array}{cc}
x_1^2 + x_2^2 + x_1 x_2,&x\in X .\\
\infty&\textrm{otherwise.}
\end{array}\right.
\end{equation*}
Now $p^{\beta} = 0$ for all $x \notin  X $, and the integral for KL distance can be reduced to an integral over $ X $. This means that samples outside $ X $ are not considered in our computations.

\subsubsection{Constant $\beta$}
\begin{figure}[h]
\begin{subfigmatrix}{2}
\subfigure[Constant $\beta$: Confidence ellipsoids]{\includegraphics[width=2.5in]{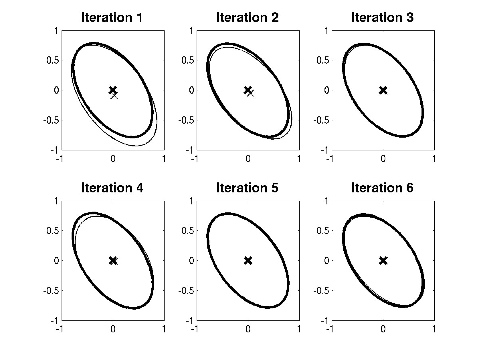}}
\subfigure[Constant $\beta$: KL distance and expected $G$]{\includegraphics[width=2.5in]{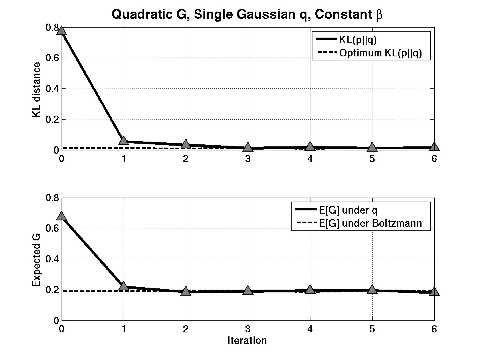}}
\subfigure[Varying $\beta$: Confidence ellipsoids]{\includegraphics[width=2.5in]{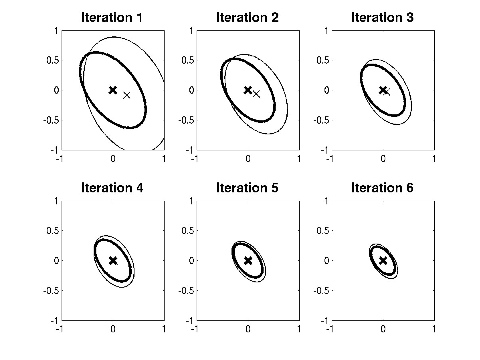}}
\subfigure[Varying $\beta$: KL distance and expected $G$]{\includegraphics[width=2.5in]{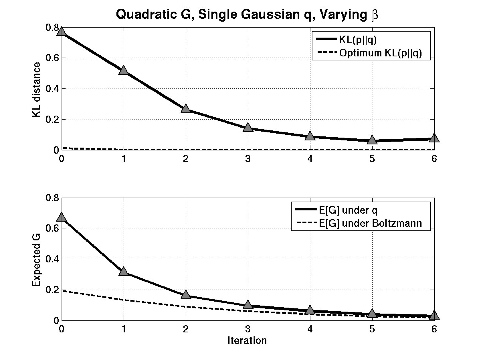}}
\end{subfigmatrix}
\caption{PC iterations for quadratic $G(x)$.}
\label{fig:quadraticGHistory}
\end{figure}

First, we fix $\beta = 5$, and run a few iterations of the PC algorithm. To start with, we draw $N_j = 30$ samples from the uniform distribution on $ X $. The best-fit Gaussian is computed using the immediate sampling procedure outlined in the preceding section. At each successive iteration, $N_j = 30$ more samples are drawn from the current $q_{\theta}$ and the algorithm proceeds. A total of $6$ such iterations are performed. The $90\%$ confidence ellipsoids corresponding to $p^{\beta}$ (heavy line) and the iterates of $q_{\theta}$ (thin line) are shown in Fig.~\ref{fig:quadraticGHistory}. Also shown are the corresponding values of $\mathbb{E}_{q_{\theta}} G(x)$, computed using the sample mean of $G_Q(x)$ for $1000$ samples of $x$ drawn from each $q_{\theta}$, and KL$(p^{\beta}\|q_{\theta})$, computed as the sample mean of $\ln(p^{\beta}(x)/q_{\theta}(x))$ for $1000$ samples of $x$ drawn according to $p^{\beta}$.

\subsubsection{Varying $\beta$}
Next, we change $\beta$ between iterations, in the `update $\beta$' step shown in algorithm(\ref{algo:GaussMixOverview}). With the same algorithm parameters, we start with $\beta = 10$, and at each iteration, we use a multiplicative update rule $\beta \leftarrow k_{\beta}\beta$, for some constant $k_{\beta} > 1$, in this case, $1.5$. As the algorithm progresses, the increasing $\beta$ causes the target density $p^{\beta}$ to place increasing probability mass on regions with low $G(x)$, as shown in  Fig.~\ref{fig:QuadraticSurface}. Since the distributions $q_{\theta}$ are best-fits to $p$, successive iterations will generate lower $\mathbb{E}_{q_{\theta}} G(x)$. The $90\%$ confidence ellipsoids and evolution of $\mathbb{E}_{q_{\theta}} G(x)$ and KL distance are shown in Fig.~\ref{fig:quadraticGHistory}.

\subsubsection{Cross-validation to Schedule $\beta$}
In more complex problems, it may be difficult to find a good value for the $\beta$ update ratio $k_{\beta}$. However, we note that the objective KL$(p^{\beta}\|q_{\theta})$ can be viewed as a regularized version of the original objective, $\mathbb{E}_{q_{\theta}}[G(x)]$. Therefore, we use the standard PL technique of cross-validation to pick the regularization parameter $\beta$ from some set $\{\beta\}$. At each iteration, we partition the data set $D$ into training and test data sets $D_T$ and $D_V$. Then, for each $\beta \in \{\beta\}$, we find the optimal parameters $\theta^{\star}(\beta)$ using only the training data $D_T$. Next, we test the associated $q_{\theta^{\star}(\beta)}$ on the test data $D_V$ using the following performance measure.
\begin{equation}
\label{eq:xvFitFunction}
\widehat{g}(\theta) = \displaystyle\frac{\displaystyle\sum_{D_V} \frac{q_{\theta}(x^i) G(x^i)}{h(x^i)}}{\displaystyle\sum_{D_V} \frac{q_{\theta}(x^i)}{h(x^i)}},
\end{equation}
The objective $\widehat{g}(\theta)$ is an estimate\footnote{The reason for dividing by the sum of $q(x^i)/h(x^i)$ is as follows. If the training data is such that no probability mass is placed on the test data, the numerator of $\widehat{g}_{q_{\theta}}$ is 0, regardless of the parameters of $q_{\theta}$. In order to avoid this peculiar problem, we divide by the sum of $q(x^i)/h(x^i)$, as desribed by ~\citet{roca04}.} of the unregularized objective $\mathbb{E}_{q_{\theta}}[G(x)]$. Finally, we set  $\beta^{\star} = \arg\min_{\beta\in\{\beta\}}\widehat{g}(\theta^{\star}(\beta))$, and compute $\theta^{\star}(\beta^{\star})$ using \emph{all} the data $D$. Note that the whole cross-validation procedure is carried out without any more calls to the oracle $\mathscr{G}$.

\begin{figure}[h]
\begin{subfigmatrix}{2}
\subfigure[$\mathbb{E}_q G(x)$ history.]{\includegraphics[height = 3in]{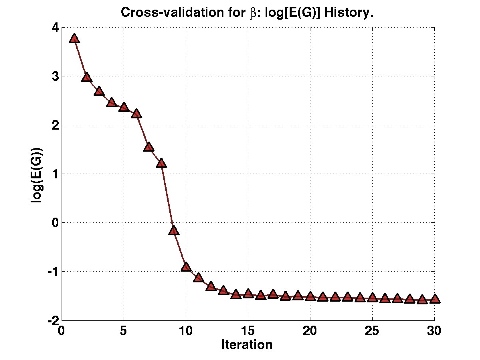}}
\subfigure[$\beta$ history.]{\includegraphics[height = 3in]{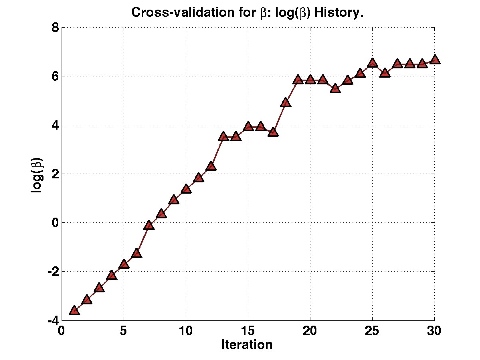}}
\subfigure[Fit $Q(\beta)$, iteration 8.]{\includegraphics[height = 3in]{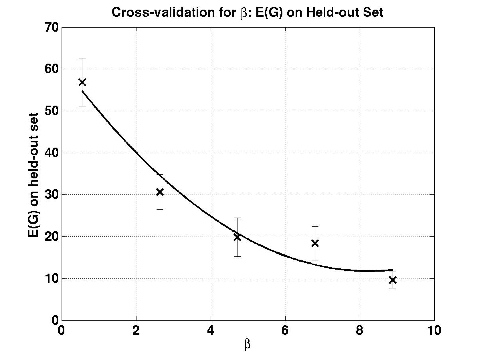}}
\subfigure[Fit $Q(\beta)$, iteration 15.]{\includegraphics[height = 3in]{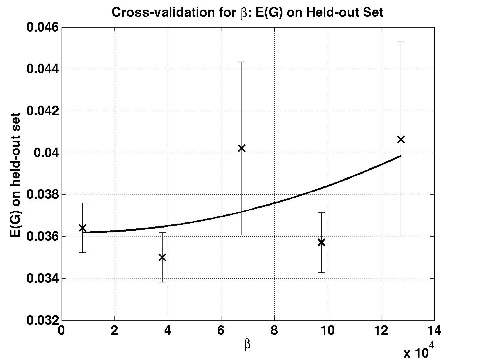}}
\end{subfigmatrix}
\caption{Cross-validation for $\beta$: 2-D Rosenbrock $G(x)$.}
\label{xvForBetaHistory}
\end{figure}

We demonstrate the functioning of cross-validation on the well-known Rosenbrock problem in two dimensions, given by
\begin{equation*}
G_R(x) = 100(x_2 - x_1^2)^2 + (1 - x_1)^2,
\end{equation*}
over the region $ X  = \{x \in \mathbb{R}^2 \;|\;\|x \|_{\infty} < 4\}$. The optimum value of 0 is achieved at $x = (1,1)$. The details of the cross-validation algorithm used are presented in Algorithm~\ref{algo:xvForBeta}.
\begin{algorithm}[h]
\caption{Cross-validation for $\beta$.}
\label{algo:xvForBeta}
\begin{algorithmic}
\STATE{Initialize interval extension count $\mathtt{extIter} = 0$, and $\mathtt{maxExtIter}$ and $\beta_0$.}.
\REPEAT
\STATE{At $\beta = \beta_0$, consider the interval $\Delta \beta = [k_1 \beta_0, k_2 \beta_0]$.}
\STATE{Choose $\{\beta\}$ be a set of $n_{\beta}$ equally-spaced points in $\Delta \beta$.}
\STATE{Partition the data into $K$ random disjoint subsets.}
\FOR{each fold $k$,}
\STATE{Training data is the union of all but the $k^{th}$ data partitions.}
\STATE{Test data is the $k^{th}$ partition.}
\FOR {$\beta_i$ in $\{\beta\}$,}
\STATE{Use training data to compute optimal parameters $\theta^{\star}(\beta_i,D_{T_k})$.}
\STATE{Use test data to compute held-out performance $\widehat{g}(\theta^{\star}(\beta_i,D_{V_k}))$, from Eq.~\ref{eq:xvFitFunction}.}
\ENDFOR
\ENDFOR
\STATE{Compute average held-out performance, $\overline{g}(\beta_i)$, of $\widehat{g}(\theta^{\star}(\beta_i,D_{V_k}))$.}
\STATE{Fit a quadratic $Q(\beta)$ in a least-squares sense to the data $(\beta_i, \overline{g}(\beta_i))$.}
\IF {$Q$ is convex}
\STATE{Set optimum regularization parameter $\beta^{\star} = \arg\min_{\beta \in \Delta\beta} Q(\beta)$.}
\ELSE 
\STATE{Fit a line $L(\beta)$ in a least-squares sense to the data $(\beta_i, \overline{g}(\beta_i))$.}
\STATE{Choose $\beta^{\star} = \arg\min_{\beta \in \Delta\beta} L(\beta)$.}
\ENDIF
\STATE{Increment $\mathtt{extIter}$}
\STATE{Update $\beta_0 \leftarrow \beta^{\star}$}
\UNTIL{$\mathtt{extIter} > \mathtt{maxExtIter}$ or $Q$ is convex.}
\end{algorithmic}
\end{algorithm}
For this experiment, we choose 
\newline\newline
\begin{tabular}{rlrlrl}
$\mathtt{maxExtIter} = $& 4, &$k_1 = $& 0.5,&$k_2 = $& 2,\\
$N_j = $& 10, &$n_{\beta} = $& 5,& $K = $& 10.\\
\end{tabular}
\newline\newline 
The histories of $\mathbb{E}_q G(x)$ and $\beta$ are shown in Fig.~\ref{xvForBetaHistory}. Also shown are plots of the fitted $Q(\beta)$ at iterations 8 and 15.
As can be seen, the value of $\beta$ sometimes \emph{decreases} from one iteration to the next, which can never happen in any fixed multiplicative update scheme.

\begin{figure}[h]
\begin{subfigmatrix}{2}
\subfigure[Least-squares fit to $\beta$:]{\includegraphics[height = 3in]{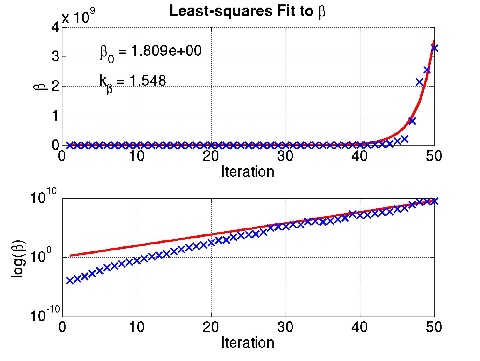}}
\subfigure[Least-squares fit to $\log(\beta)$:]{\includegraphics[height = 3in]{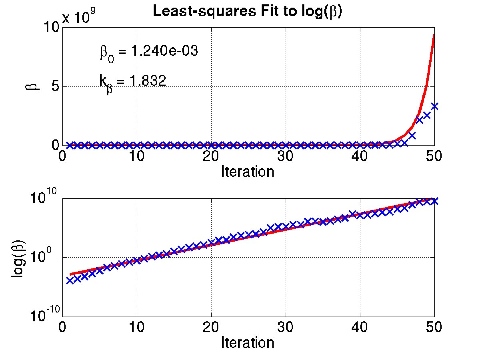}}
\end{subfigmatrix}
\caption{Best-fit $\beta$ update rule.}
\label{bestFitBeta}
\end{figure}

\begin{figure}[h]
\begin{subfigmatrix}{2}
\subfigure[$\log(\mathbb{E}_q G)$ history.]{\includegraphics[height = 3in]{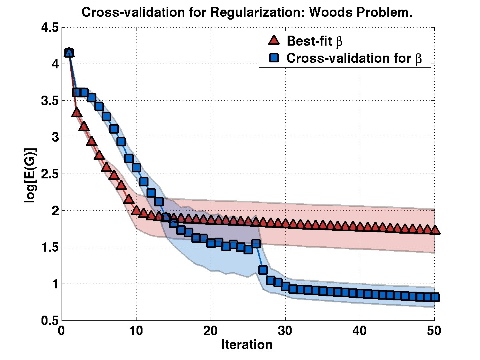}}
\end{subfigmatrix}
\caption{Cross-validation beats best-fit fixed $\beta$ update: 4-D Woods $G(x)$.}
\label{fig: xvForBetaWoods}
\end{figure}

We now demonstrate the need for an automated regularization scheme, on another well-known test problem in $\mathbb{R}^4$, the Woods problem, given by
\begin{eqnarray*}
G_{\mathrm{woods}}(x) &=& 100(x_2 - x_1)^2 + (1 - x_1)^2 + 90(x_4 - x_3^2)^2 + (1 - x_3)^2  \nonumber\\
     &&+ 10.1[(1 - x_2)^2 + (1 - x_4)^2] + 19.8(1-x_2)(1 - x_4).
\end{eqnarray*}
The optimum value of 0 is achieved at $x = (1,1,1,1)$. We run the PC algorithm 50 times with cross-validation for regularization. For this experiment, we used a single Gaussian $q$, and set 
\newline\newline
\begin{tabular}{rlrlrl}
$\mathtt{maxExtIter} = $& 4,&$k_1 = $& 0.5,&$k_2 = $& 3,\\
$N_j = $&20,&$n_{\beta} = $& 5,& $K = $& 10.\\
\end{tabular}
\newline\newline 
From these results, we then attempt to find the best-fit multiplicative update rule for $\beta$, only to find that the average $\beta$ variation is not at all well-approximated by \emph{any} fixed update $\beta \leftarrow k_{\beta} \beta$. This poor fit is shown in Fig.~\ref{bestFitBeta}, where we show a least-squares fit to both $\beta$ and $\log(\beta)$. In the fit to $\log(\beta)$ the final $\beta$ is off by over 100\%, and in the fit to $\beta$, the initial $\beta$ is off by several orders of magnitude.
We then compare the performance of cross-validation to that of PC algorithm using the fixed update rule derived from the best least-squares fit to $\log(\beta)$. From a comparison over 50 runs, we see that using this best-fit update rule performs extremely poorly - cross-validation yields an improvement in final $\mathbb{E}_{q_{\theta}}G(x)$ by over an order of magnitude, as shown in Fig.~\ref{fig: xvForBetaWoods}.

\subsubsection{Bagging}
While regularization is a method to decrease bias, bagging is a well-known variance-reducing technique. Bagging is easily incorporated in our algorithm. Suppose, at some stage in the algorithm, we have $N$ samples $(x^i,g^i)$, we resample our existing data set exactly $N$ times with replacement. This gives us a different set of data set $D'$, which also contains some duplicates. We compute optimal parameters $\theta^{\star}(D')$. We repeat this resampling process $k_b$ times and uniformly average the resulting optimal densities $q_{\theta^{\star}(D'_k)}, k=1,\ldots,k_b$. 

\begin{figure}[h]
\centering\includegraphics[width=3in]{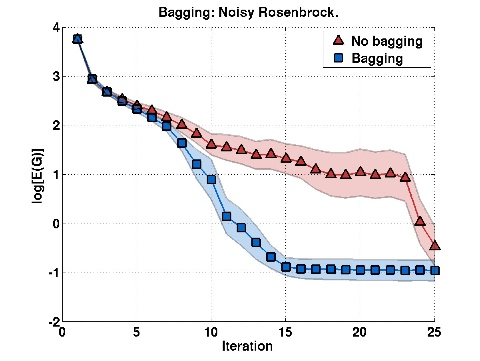}
\caption{Bagging improves performance: Noisy 2-D Rosenbrock.}
\label{baggingHistory}
\end{figure}

We demonstrate this procedure, using the Rosenbrock function and a single Gaussian $q_{\theta}$. In this experiment, we also demonstrate the ability of PC to handle non-deterministic oracles by adding uniform random noise to every function evaluation, that is, $(g\mid x,G) \sim \mathcal{U}[-0.25,0.25]$. For this experiment, $N_j = 20, k_b = 5$. The $\beta$ update is performed using the same cross-validation algorithm described above. Fig.~\ref{baggingHistory} shows the results of 50 runs of the PC algorithm with and without bagging. 
We see that bagging finds better solutions, and moreover, it reduces the variance between runs. Note that the way we use bagging, we are only assured of improved variance for a single MC estimation at a given $\theta$, and not over the whole MCO process of searching over $\theta$.

\subsubsection{Cross-validation for Regularization and Model Selection}
\begin{figure}[h]
\centering\includegraphics[width=3in]{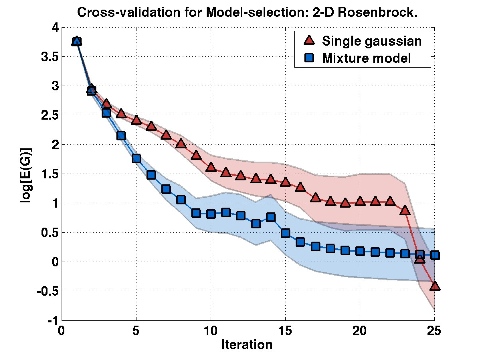}
\caption{Cross-validation for regularization and model-selection: 2-D penalty function $G(x)$.}
\label{mixturesHistory}
\end{figure}

In many problems like the Rosenbrock, a single Gaussian is a poor fit to $p^{\beta}$ for many values of $\beta$.
In these cases, we can use a mixture of Gaussians to obtain a better fit to $p^{\beta}$. We now describe the use of cross-validation to pick the number of components in the mixture model. We use an algorithm very similar to the one described for regularization. In these experiments, we use a greedy algorithm to search over the joint space of $\beta$ and models: 
\begin{enumerate}
\item
We first pick the regularization parameter $\beta$, using Algorithm~\ref{algo:xvForBeta}.
\item 
For that $\beta$, we use Algorithm~\ref{algo:xvForModel}Ä to pick the number of mixture components.
\end{enumerate}
\begin{algorithm}
\caption{Cross-validation for model selection.}
\label{algo:xvForModel}
\begin{algorithmic}
\STATE{Initialize set $\{\{\phi\}\}$ of model classes $\{\phi\}$ to search over.}
\STATE{Partition the data into $K$ disjoint subsets.}
\FOR{each fold $k$,}
\STATE{Training data is all but the $k^{th}$ data partitions.}
\STATE{Test data is the $k^{th}$ data partition.}
\FOR {$\{\phi_i\}$ in $\{\{\phi\}\}$}
\STATE{Compute the optimal parameter set $\theta^{\star}(D_{T_k}) \in \{\phi_i\}$}
\STATE{Compute held-out performance $\widehat{g}(\theta^{\star}(D_{V_k}))$}
\ENDFOR
\STATE{Compute the sample held-out performance, $\overline{g}(\{\phi_i\})$, from Eq.~\ref{eq:xvFitFunction}.}
\ENDFOR
\STATE{Choose best model class $\{\phi^{\star}\} = \arg\min_{\phi_i} \overline{g}(\{\phi_i\})$.}
\end{algorithmic}
\end{algorithm}
For this experiment, the details are the same as the preceding section, but without bagging. The set of models $\{\{\phi\}\}$ is the set of Gaussian mixtures with one, two or three mixing components. Fig.~\ref{mixturesHistory} shows the variation of $\mathbb{E}_q(G)$ vs. iteration.
The mixture model is much quicker to yield lower expected G, because the Boltzmann at many values of $\beta$ is better approximated by a mixture of Gaussians. However, note that the mixture models performs poorly towards the end of the run. The reason for this is as follows: No shape regularization was used during the EM procedure. This means that the algorithm often samples from nearly degenerate Gaussians. These `strange' sample sets hurt the subsequent performance of importance sampling, and hence of the associated MCO problem. This can be alleviated by using some form of shape regularization in the EM algorithm.

\section{Fit-based Monte Carlo}
\label{sec:fit_based}
Thus far, we have not exploited the locations of the samples in constructing esimates. In this section, we discuss the incorporation of sample locations to improve both MC and MCO.

\subsection{Fit-based MC Estimation of Integrals}

We first consider MC estimation of an integral,
presented at the beginning of Sec.~\ref{sec:covars}. Recall from that
discussion, that to accord with MCO notation, we write the integral to
be estimated as ${\mathscr{L}}(\phi) = \int dw \; U(w, \phi)$ for some
fixed $\phi$. In this notation the sampling of $v$ provides a sample
set $\{(w^i, U(w^i, \phi)) : i = 1, \ldots N)$. The associated sum
${\hat{\mathscr{L}}}_{\{(w^i, U(w^i, \phi))\}} \equiv {\hat{\mathscr{L}}}(\phi)$ then serves as our
estimate of the integral ${\mathscr{L}} \equiv {\mathscr{L}}(\phi)$.

In forming the estimate ${\hat{\mathscr{L}}}_{\{(w^i, U(w^i,
\phi))\}}$ we do not exploit the relationships between the
locations of the sample points and the associated values of the
integrand. Indeed, since those locations \{$w^i$\} do not appear directly in
that estimate, ${\hat{\mathscr{L}}}_{\{(w^i, U(w^i, \phi))\}}$ is unchanged even 
if those locations changed in such a way that the values \{$r^i$\} stayed the same.

The idea behind {\bf{fit-based}} (FB) Monte Carlo is to leverage the
location data to replace ${\hat{\mathscr{L}}}_{\{(w^i, U(w^i, \phi))\}}$ with a more accurate
estimate of $\mathscr{L}$.  The most straightforward FB MC treats the
sample pairs \{$(w^i, U(w^i, \phi)) : i = 1, \ldots N)$\} as a
training set for a supervised-learning algorithm. Running such an
algorithm produces a fit ${\tilde{U}}(., \phi)$ taking $w$'s into
$\mathbb{R}$. This fit is an estimate of the actual oracle $U(,
\phi)$, 
and this fit defines an estimate for the full integral,
\begin{eqnarray}
{\tilde{\mathscr{L}}}_{\{w^i, U(w^i, \phi)\}} \equiv \int dw \; {\tilde{U}}(w, \phi) 
\label{eq:hat_I}
\end{eqnarray}
We will sometimes omit the subscript and just
write ${\tilde{\mathscr{L}}}(\phi)$ or even just
${\tilde{\mathscr{L}}}$.  In this most straightforward version of FB
MC, we use ${\tilde{\mathscr{L}}}$ as our estimate of $\mathscr{L}$
rather than ${\hat{\mathscr{L}}}$.

In some circumstances one can evaluate ${\tilde{\mathscr{L}}}$ in
closed form.  A recent paper reviewing some work on how to do this
with Gaussian processes is given by~\citet{ragh03}. In other circumstances one
can form low-order approximations to ${\mathscr{L}}$, for example
using Laplace approximations~\citep[see][]{roca04}. Alternatively,
conventional deterministic grid-based approximation of the integral
${\mathscr{L}}$ can be cast as a degenerate version of closed-form
fit-based estimation{\footnote{To see this, modify
the MC process to be sampling without replacement. Choose the proposal
distribution $v$ for this process to be a sum of delta functions. The
centers of those delta functions give the points on a regular grid of
points in the space of allowed $x$'s. Have the number of samples equal
the number of such grid points. Finally, have our fit to the samples,
${\tilde{U}}(., \phi)$, be a sum of step-wise constant functions,
going through the sample points. The closed form integral of that fit
given by Eq.~\ref{eq:hat_I} is just the Reimann approximation to the
original integral, $\int dw \; U(w, \phi)$.}}
 of ${\mathscr{L}}$.
 
More generally, one can form an approximation to the integral
${\tilde{\mathscr{L}}}(\phi)$ by MC sampling of ${\tilde{U}}(.,
\phi)$. Generating these {\bf{fictitious samples}} of
$\tilde{U}(.,\phi)$ does not incur the expense of calling the actual
oracle $U(., \phi)$. So, in this approach, we run MC twice. The first
time, we generate the {\bf{factual samples}} \{$(w^i, U(w^i, \phi))
: i = 1, \ldots N)$\}. Given those samples, we form the fit to them,
$\tilde{U}(., \phi)$. We then run a second MC process using the
fictitious oracle $\tilde{U}(., \phi)$.

Note that in all of these approaches, the original sampling distribution $v$ does
not directly arise, that is, the values \{$v(w^i)$\} do not arise. In
particular, if one were to change those values without changing the
factual sample locations $\{w^i\}$, then the estimate
${\tilde{\mathscr{L}}}_{\{(w^i, U(w^i, \phi))\}}$ would not change.
Of course, a different $v$ would result in a different sample set, and thereby a different estimate, but \emph{given a sample set}, the sampling distribution is immaterial.
This is typically the case with FB MC
estimators, and it contrasts with the estimator
${\hat{\mathscr{L}}}_{\{(w^i, U(w^i, \phi))\}}$, which would change if
$v$ were changed without changing the factual sample locations.

Note also that the factual samples underlying the fit ${\tilde{U}}(.,
\phi)$ are exact samples of the factual oracle, $U(., \phi)$. In
contrast, since in general ${\tilde{U}}(., \phi) \ne U(., \phi)$, the
fictitious samples will be erroneous, if viewed as samples of $U(.,
\phi)$. Since we are ultimately concerned with an integral of $U(.,
\phi)$, this suggests that the fictitious samples should be weighted
less than the factual samples. This might be the
case even if one had infinitely many fictitious samples. In fact, even
if one could evaluate ${\tilde{\mathscr{L}}}$ in closed form, it might
make sense not to use it directly as our final estimate of
${\mathscr{L}}$. Instead, combining it with the importance-sampling
estimate ${\hat{\mathscr{L}}}$ might improve the estimate.

\subsection{Bayesian Fit-based MCO}

We now extend the discussion to MCO by allowing $\phi$ to vary. As
with the case of FB MC estimation of an integral, the most
straightforward version of FB MCO uses ${\tilde{\mathscr{L}}}(\phi)$
rather than ${\hat{\mathscr{L}}}(\phi)$ as our estimate of
${\mathscr{L}}(\phi)$.  This means that we use argmin$_\phi
[{\tilde{\mathscr{L}}}(\phi)]$ rather than argmin$_\phi
[{\hat{\mathscr{L}}}(\phi)]$ as our estimate of the $\phi$
optimizing ${\mathscr{L}}(\phi)$.

${\hat{\mathscr{L}}}(\phi)$ is a sum, whereas
${\tilde{\mathscr{L}}}(\phi)$ is an integral. This means that
different algorithms are required to find the $\phi$ optimizing
them. Indeed, optimizing ${\tilde{\mathscr{L}}}(\phi)$ is formally the
same type of problem as optimizing ${\mathscr{L}}(\phi)$; both
functions of $\phi$ are parametrized integrals over $w$.  So if
needed, we can use PLMCO techniques to optimize
${\tilde{\mathscr{L}}}(\phi)$. Again, as with MC, the integrand of
${\tilde{\mathscr{L}}}(\phi)$ is not the factual oracle. So, minimizing ${\tilde{\mathscr{L}}}(\phi)$ using PLMCO sampling techniques will not require making additional calls to the factual oracle.

Consider a Bayesian approach to
forming the fit.  Our problem is to solve the MCO problem (P2), given
that the factual oracle $U$ is not known, and we only can generate samples
of $U$.  Adopting a fully Bayesian
perspective, since $U$ is not known, we must treat it as a random
variable. So we have a posterior distribution over all possible
oracles, reflecting all the data we have concerning the factual
oracle.  We then use that posterior to try to solve (P2).

Say that in the usual way that our data contains the $w$'s and the
associated functions $U(w, .)$ of a sample set that was generated by
importance sampling $U$ (see Sec.~\ref{sec:sup_equals_mco}).  More
generally, we may have additional data, for instance, the gradients of $U$ at the sample points. For
simplicity though, we restrict attention to the case where the
provided information is only the sample set of functions, $\{w^i,
r^i(.)\}$, together with $v$.

We will use $D$ to refer to a sample set for MC or MCO, and for
immediate sampling in particular.  So we write our posterior over
oracles{\footnote{Practically, when running a
computer experiment, $U$ is the actual oracle generating $D$ according
to a likelihood $P(D \mid U$). On the other hand, the posterior $P(U_c
\mid D)$ reflects both that likelihood and a prior $P(U_c)$ assumed by the
algorithm. So, $U_c$ is a random variable, whereas
$U$ refers to the single true factual oracle.}} as $P(U_c \mid D)$. Using this notation, the
goal in Bayesian FB MCO 
is to exploit $P(U_c \mid D)$ to improve our estimate of the $\phi$ that minimizes $\int dw
\; U(w, \phi)$.

How should we use $P({U_c} \mid D)$ to estimate the solution to (P2)?
Bayesian decision theory tells us to minimize posterior expected loss,
$\int d{U_c} \; P({U_c} \mid D) L(\phi, {U_c})$.
Given the loss function of Eq.~\ref{eq:mco_loss}, that means we wish to solve
\begin{eqnarray*}
\textrm{(P6): }\min_{\phi}
\int d{U_c} \; P({U_c} \mid D) L(\phi, {U_c})
&=& \min_{\phi} \int dw_cd{U_c} \; P({U_c} \mid D) {U_c}(w_c, \phi) .
\end{eqnarray*}
To avoid confusion, the variable of integration is written as $w_c$, to
distinguish it from $w$'s in the integral $\int dw \; U(w, \phi)$.
The solution to (P6) is our best possible guess of the $\phi$ solving
problem (P2), given the sample set $D$.  Finding that solution is a
problem of minimizing a parametrized integral.

Sometimes we may be able to solve (P6) in closed form, even when we
cannot solve (P2) in closed form. Performing the integral over
${U_c}$ may simplify the remaining integral over $w_c$. More generally,
we can address (P6) using MCO techniques, and in particular using
PLMCO.{\footnote{To see that explicitly, rewrite the integral in
(P2) as $\int dz
\; V(z, \phi)$, and identify values of $z$ with pairs $(w_c, {U_c})$, while 
taking $V(z, \phi) = V(w_c, {U_c}, \phi) = P({U_c} \mid D) {U_c}(w_c, \phi)$.}}  

To solve (P6) with PLMCO one generates fictitious samples by sampling
one distribution over ${U_c}$'s and one over $w_c$'s. 
This MC process does not involve calls to the actual oracle $U$, but samples a new distribution
over ${U_c}$'s, to generate counter-factual ${U_c}$'s, and then samples those
${U_c}$'s. 

\subsection{Example: Fit-based Immediate Sampling} 

To illustrate the foregoing we consider the variant of MCO given by
immediate sampling with a noise-free oracle.  In the simple version of
MCO considered just above, the estimate we make for $\phi$ has no
effect on what points are chosen for any future calls we might make to
the oracle. For simplicity, we restrict attention to the analogous
formulation of immediate sampling.  Using immediate sampling terminology, 
this means that we only consider
the issue of how best to estimate $\theta$ after the immediate
sampling algorithm has exhausted all its calls to the oracle.  We do
not consider the more general active learning issue, of how best to
estimate $\theta$ when this estimate will be affect further calls to the oracle.
However see Sec.~\ref{sec:exp_loss_mco} below.

Recall that in immediate sampling, identifying $w_c$ with $x_c$ and
$\phi$ with $\theta$, $U_c(w_c, \phi)$ becomes $F(G_c(x_c), \theta)$.
Given a sample set $D$ of $(x,G(x))$ pairs generated from a noise-free
factual oracle, our Bayesian optimization problem in immediate
sampling is  to find the $\theta$ that
minimizes
\begin{eqnarray}
{\mathbb{E}}({\mathscr{F}}_{G_c}(\theta) \mid D) &=& 
{\mathbb{E}}_{P(G_c \mid D)} [\int dx_c F(G_c(x_c), q_\theta(x_c))] \nonumber \\ 
&=& \int dx_c dG_c \; P(G_c \mid D) F(G_c(x_c), q_\theta(x_c)) \nonumber \\
&\triangleq& {\tilde{\mathscr{F}}}_D(\theta).
\label{eq:bayes_imm_single_valued}
\end{eqnarray}
Contrast this with Eq.~\ref{eq:expected_G}. In particular, the inner integral in
Eq.~\ref{eq:bayes_imm_single_valued} runs over fictitious oracles
${G_c}$ that are generated according to $P({G_c} \mid D)$, whereas in 
Eq.~\ref{eq:expected_G}, $G$ is the factual oracle.

In some circumstances one can evaluate the integral in
Eq.~\ref{eq:bayes_imm_single_valued} algebraically, to give a closed
form function of $\theta$. In other cases, we can algebraically
evaluate an accurate low-order approximation, to again give a closed
form function of $\theta$. For the rest of this subsection, however, we consider
the situation where neither of these possibilities hold.

To address this situation, we approximate
the integral in Eq.~\ref{eq:bayes_imm_single_valued} using importance
sampling. However, to do this, we may have to
importance-sample over two domains. The first sampling is over sample
locations $x_c$, using some sampling distribution $h_c$. (As an
example, we can simply choose $h_c = h$.) The second sampling is over
possible oracles $G_c$, using some sampling distribution $H$.  More
precisely, write
\begin{eqnarray}
{\mathbb{E}}({\mathscr{F}}_{G_c}(\theta) \mid D) &=& \int dx_c\,d{G_c}
\; \bigl[h_c(x_c) H({G_c}) \bigr] \; 
\frac{P({G_c} \mid D)}{h_c(x_c) H({G_c})} F({G_c}(x_c), q_\theta(x_c)) .
\end{eqnarray}
To approximate this integral generate $N_T$ locations, \{$x_c^i$\}, by
sampling $h_c$. This gives us $N_T$ integrals
\begin{eqnarray}
T_{D,\{x_c^i\}}(\theta) &\triangleq& \frac{1}{{h_c}({x_c^i})} \int dG_c \; H(G_c) \frac{P({G_c} \mid D)}{H({G_c})}
F({G_c}({x_c^i}), q_\theta({x_c^i})) .
\end{eqnarray}
Sometimes, these integrals also can be evaluated algebraically, giving
closed form sample functions of $\theta$. As an example, suppose we 
sample oracles according to the posterior, that is, take $H(G_c)
= P(G_c \mid D)$, so that 
\begin{eqnarray}
T_{D,\{x_c^i\}}(\theta) = \frac{1}{{h_c}({x_c^i})} \int dG_c \; P({G_c}
\mid D) F({G_c}({x_c^i}), q_\theta({x_c^i})) . 
\end{eqnarray}
Next, say that we have a Gaussian process prior over oracles,
$P(G_c)$~\citep{rawi06}, with a Gaussian covariance
kernel. For this choice, and for some $F$'s, we can compute
$T_{D,\{x_c^i\}}(\theta)$ exactly for any $\theta$, provided $D$ is not
too large. For example, this is the case for most $F$'s whose
dependence on their first argument lies in the exponential
family.{\footnote{Strictly speaking, since the oracle is noise-free,
the likelihood $P(D \mid G_c)$ is a delta function about having $D$
lie exactly on the function $G_c(x)$. In practice, this may make the
computation be ill-behaved numerically. Typically such problems are
addressed modeling the fictitious oracles as though the values they
returned had a small amount of Gaussian noise added.}} In other
situations, while we cannot evaluate them exactly, the integrals
$T_{D,\{x_c^i\}}(\theta)$ can be accurately approximated
algebraically. This again reduces them to closed form sample functions
of $\theta$.

In either of these cases, there is no need to sample $H$; samples of
$h_c$ suffice. More generally, we may not be able to evaluate the $N_T$
functions $T_{D,\{x_c^i\}}(\theta)$ algebraically, and also cannot form
accurate low-order algebraic approximations to them. In this
situation, for each ${x_c^i}$, we should generate one sample of ${G_c}$
from $H({G_c})$.{\footnote{ One could have the number of samples of
$H({G_c})$ not match the number of samples of $h_c(x)$. To avoid the
associated notational overhead, here we just match up the two types of
samples, one-to-one.}} This provides us a total of $N_T$ sample
functions
\begin{eqnarray}
T_{D,\{x_c^i,G^i_c\}}(\theta) &\triangleq& \frac{ P(G^i_c \mid D) F(G^i_c({x_c^i}),
q_\theta({x_c^i}) )}{{h_c}({x_c^i})H(G_c^i)}.
\label{eq:fb_samp_func}
\end{eqnarray}

As an example, say we believe firmly that a particular posterior
$P({G_c} \mid D)$ governs our problem, and can sample that
posterior. Then we can take $H({G_c})$ to equal $P({G_c} \mid D)$. Now
Eq.~\ref{eq:fb_samp_func} only requires that we have values of our
sampled ${G_c}$'s at the $\{{x_c^i}\}$, that is, we only need to have
values $G^i_c({x_c^i})$, where $G^i_c \sim H({G_c})$. So we only need
to sample the $N_T$ separate one-dimensional distributions
\{$P({G_c}(x_c^i) \mid D)$\}. In particular, we might be
able to use Gaussian Process techniques to generate those values.
Alternatively, as a rough approximation, one could simply fit a
regression to the data in $D$, $\omega(x)$, and then add noise to the
vector of values \{$\omega({x_c^i})$\} to get
\{$G^i_c({x_c^i})$\}. If we wanted to have multiple $G_c$'s for each
$x_c^i$, then we would simply generate more samples of each
distribution \{$P({G_c}(x_c^i) \mid D)$\}.{\footnote{As an
alternative, we could reverse the sampling order and sample $P(G_c
\mid D)$ first and $h_c$ second. Practically, this would mean
generating $N_T$ samples of $h_c$, and then sampling the single
$N_T$-dimensional distribution $P(G_c(x_c^1), \ldots G_c(x_c^{N_T})
\mid D)$. (This contrasts to the case considered in the text in which $H$ is sampled second,
so one instead samples $N_T$ separate one-dimensional distributions
\{$P({G_c}(x_c^i) \mid D)$\}.) If we do this using a `rough
approximation' based on a fit $\omega$ to $D$, the noise values added
to the values of the fit, $\{\omega(x_c^i)\}$, would have to be
correlated with each other, since they reflect the same $G_c$.}}

However we sample $H$, the resultant sample functions provide the
estimate
\begin{eqnarray}
{\hat{\tilde{\mathscr{F}}}}_{D,\{{x_c^i},G^i_c\}}(\theta) 
&\triangleq& \frac{1}{N_T}\sum_{i = 1}^{N_T} T_{D,\{{x_c^i},G^i_c\}}(\theta) \nonumber \\
&\approx& {\mathbb{E}}({\tilde{\mathscr{F}}}_{D, \{{x_c^i},G^i_c\}}(\theta) \mid D) 
\label{eq:bayes_imm_sum}
\end{eqnarray}
which we sometimes abbreviate as ${\hat{\tilde{\mathscr{F}}}}_{D}(\theta)$.
For the situation where we can evaluate the integral over $G_c$
algebraically (or at least approximate it that way), we instead define
\begin{eqnarray}
{\hat{\tilde{\mathscr{F}}}}_{D}(\theta) &\triangleq& \frac{1}{{N_T}}\sum_{i =
1}^{N_T} T_{D,{\{x_c^i\}}}(\theta)
\label{eq:bayes_alg_imm_sum}
\end{eqnarray}
and rely on the context to decide which of the definitions of
${\hat{\tilde{\mathscr{F}}}}_{D}(\theta)$ is meant. So
for either case we can write ${\tilde{\mathscr{F}}}_{D}(\theta)
\approx {\hat{\tilde{\mathscr{F}}}}_{D}(\theta)$.{\footnote{As a
practical issue, we may want to divide the sum in
Eq.~\ref{eq:bayes_alg_imm_sum} by the empirical average $\sum_{i =
1}^{N_T} \frac{P(G^i_c \mid D)} {h_c(x_c^i)}$. Similarly, if we cannot
evaluate the integral over $G_c$ algebraically, we may want to modify
the estimate in Eq.~\ref{eq:bayes_imm_sum} by dividing by $\sum_{i =
1}^{N_T} \frac{P(G^i_c \mid D)} {H(G^i_c){h_c}({x_c^i})}$. Such divisions would accord
with the analogous division we do in our non-FB immediate sampling
experiments.}}

In both cases, under naive MCO, we search for the $\theta$ that
minimizes ${\hat{\tilde{\mathscr{F}}}}_{D}(\theta)$. We then use that
$\theta$ as our estimate for the solution to (P6).  More generally,
rather than use naive MCO we can exploit our sample functions with
PLMCO. For example, rather than minimizing
${\hat{\tilde{\mathscr{F}}}}_{D}(\theta)$, we could minimize a sum of
${\{\tilde{\mathscr{F}}}_{D}(\theta)$ and a regularization penalty term.

However we arrive at our (estimated) optimal $q_{\theta}$, most
simplistically, we can update $h$ to equal that new $q_{\theta}$. In a
more sophisticated approach, we could set $h$ from the sample
functions using active learning (see Sec.~\ref{sec:exp_loss_mco}
below). Once we have that new $h$ we can form samples of it to
generate new factual sample locations $x$. These in turn are fed to
the factual oracle $G$ to augment our data set $D$. Then the process
repeats.

Note that unlike with non-FB immediate sampling, with FB immediate
sampling we need to evaluate $P({G_c} \mid D)$ (or sample it, if we
choose to have $H({G_c})
\triangleq P({G_c} \mid D)$). This may be non-trivial. On
the other hand, that very same distribution $P({G_c} \mid D)$ that may
cause difficulty also gives the major advantage of the fit-based
approach; it allows any insights we have into how to fit a curve
${G_c}$ to the data points $D$ to be exploited.

\subsection{Exploiting FB Immediate Sampling}

To illustrate how fitting might improve immediate sampling, consider
the case where ${\mathscr{F}}_G(q_\theta)$ is $qp$ KL distance. Say
that $G(x)$ is a high-dimensional convex paraboloid inside a
hypercube, and zero outside of that hypercube. Suppose as well that we
have a single factual sampling distribution $h$, which is concentrated
on one side of the paraboloid. For example, if the peak of the
paraboloid is at the origin, $h$ might be a Gaussian (masked by the
hypercube) whose mean lies several sigmas away from the origin.

To start, consider importance sampling MC estimation of the integral
${\mathscr{F}}_G(q_\theta)$ for one particular $\theta$, without any
concern about choosing among $\theta$'s. Say that the factual sample
$D$ isn't too large. Then it is likely that no elements of $D$ are in
regions where $G$ reaches its lowest values. For such a $D$, the
associated factual estimate
\begin{eqnarray}
{\hat{\mathscr{F}}}_{D}(\theta) &\triangleq& \frac{\sum_i
r^i_{h}(x^i, \theta)}{N}
\end{eqnarray}
is larger than the actual value, ${\mathscr{F}}_G(q_\theta)$
(cf. Eq.~\ref{eq:max_like}). So straightforward importance-sampling
integral estimation is likely to be badly off.{\footnote{Since
importance sampling is unbiased, this means its variance is likely to
be large.}}

Intuitively, the problem is that as far as the factual estimate
${\hat{\mathscr{F}}}_{D}(\theta)$ is concerned, $G$ could just as well
be a sum of delta functions centered at the $x$'s in $D$, with low
associated oracle sample values, as a paraboloid. If $G$ were in fact
such a sum, then ${\hat{\mathscr{F}}}_{D}$ would be correct. However
by looking at the $(x, G(x))$ pairs in $D$, all of which lie on the
same paraboloid, such an inference of $G$ appears quite
unreasonable. It makes sense to instead infer that $G$ is a paraboloid.

Fitting is a way to formalize (and exploit) such $D$-based
insights. As an example, consider using a Bayesian PL algorithm to do
the fitting.  Typical choices for the prior $P({G_c})$ used in PL
would result in a posterior $P({G_c} \mid {D})$ that would be far more
tightly concentrated about the actual $G$'s paraboloid shape than
about the sum of delta functions. Fitting would automatically reflect
this, and thereby produce a better estimate of
${\mathscr{F}}_G(\theta)$ than
${\hat{\mathscr{F}}}_{D}(\theta)$.{\footnote{It might be objected that
in a different problem $G$ actually would be the sum of delta
functions, not the paraboloid. In that case the FB estimate is the one
that would be in error. However this possibility is exactly what the
prior $P({G_c})$ addresses; if in fact you have reason to believe that
a $G$ that is a sum of delta functions is {\it{a priori}} just as
likely as a paraboloid $G$, then that should be reflected in
$P({G_c})$. Doing so would in turn make the FB estimate more closely
track the non-FB estimate.}}

Now we aren't directly concerned with the accuracy of our estimate
${\mathscr{F}}_G(q_\theta)$ for any single $\theta$. We aren't even
concerned with the overall accuracy of that estimate for a set of
$\theta$'s. Rather we are concerned with the accuracy of the ranking
of the $\theta$'s given by those estimates.  For example, consider
naive MCO, under which we choose the $\theta$ minimizing
${\tilde{\mathscr{F}}}_{D}(\theta)$. Even if all of our estimates (one
for each $\theta$) were far from the associated actual values, if
their signed errors were identical, the naive MCO would perform
perfectly. 

In other words, ultimately we are interested in correlations between
errors of our estimates of ${\mathscr{F}}_G(q_\theta)$ for different
$\theta$'s. (See Sec.~\ref{sec:covars}.)  Nonetheless, we might expect
that if we tend to have large error in our estimates of
${\mathscr{F}}_G(q_\theta)$ for the $\theta$'s, then everything else
being equal, we would be likely to have large error in the associated
estimate of an optimal $\theta$.

In Sec.~\ref{sec:impls} we exploited the equivalence between PL and MCO
to improve upon naive MCO. However the parameter in MCO doesn't
specify a functional fit to a data set.  Accordingly, the
incorporation of PL into MCO considered in Sec.~\ref{sec:impls} doesn't
involve fitting a function to $D$. This is why those PLMCO techniques
don't address the issue raised in this example; fitting does that. So
in full FB MCO, we may use PL in two separate parts of the algorithm,
both to form the fit to $D$, and then to use those fits to choose
among the $\theta$'s.

\subsection{Statistical Analysis of FB MC}
\label{sec:exp_loss}

Before analyzing expected performance of FB MCO, we start with the
simpler case of FB MC introduced at the beginning of this section. For
simplicity we assume that the integral ${\tilde{\mathscr{L}}}_D$ can
be calculated exactly for any $D$, so that no fictitious samples
arise.

As discussed in Sec.~\ref{sec:covars}, two important properties of an
MC estimator of an integral ${\mathscr{L}}(\phi) = \int dw \; U(w,
\phi)$ are the sample bias and the sample variance of that
estimator. Together, these give the expected loss of the estimator
under a quadratic loss function, conditioned on a fixed oracle $U(.,
\phi)$.

This is just as true for a Bayesian fitting algorithm as for any
other.  For quadratic loss, for sample set $D \equiv \{w^i, U(w^i,
\phi)\}$, the Bayesian FB MC prediction for ${\mathscr{L}}$ is
the posterior mean, 
\begin{eqnarray}
{\tilde{\mathscr{L}}}_D &=& \int d{w}' \; dU_c(.,
\phi) \; U_c({w}', \phi) P(U_c(., \phi) \mid D).
\end{eqnarray}
Accordingly, the expected quadratic loss of Bayesian FBMC is
\begin{eqnarray}
&&\!\!\!\!\!\!\!\! \int dD \; P(D \mid v, U(., \phi)) [{\mathscr{L}} -
{\tilde{\mathscr{L}}}_D]^2 \;\; = \nonumber \\
&& \int d{w}^1 \ldots d{w}^{N} \; \prod_{i=1}^{N} v({w}^i) [\int d{w}' \; U({w}', \phi) \;-\; 
\int d{w}' \; dU_c(, \phi) \; U_c({w}', \phi) P(U_c(., \phi) \mid D)]^2
\label{eq:bayes_samp}
\end{eqnarray}
where $v$ is the proposal distribution that is IID sampled ${N}$ times
to generate the sample set. 

In the usual way one can re-express this expected quadratic loss using
a bias-variance decomposition.  Whereas a conventional importance
sample estimator of $\int d{w} \; U(w,
\phi)$ is unbiased, the Bayesian estimator is biased in general; typically
\begin{eqnarray}
\int dD \; P(D \mid v, U(., \phi)) \; {\tilde{\mathscr{L}}}_D
&\ne& {\mathscr{L}} .
\label{eq:bayes_biased}
\end{eqnarray}
This bias is a general characteristic of Bayesian
estimators. Furthermore, for some functions $U(., \phi)$, the Bayesian
estimator will both be biased (unlike the factual sample estimator)
and have higher variance than the factual sample estimator.  So for
those $U(., \phi)$, the Bayesian estimator has worse bias plus
variance.

In conventional importance sampling estimation of an integral, the sampling
distribution $v$ is used twice. First it is used to form the sample
set. Then, when the sample set has been formed, $v$ is used again, to
set the denominator values in the ratios giving the MC estimate of the
integral (cf. Sec.~\ref{sec:mco}). In contrast, Bayesian FB MC
doesn't care what $v$ is. $P(U_c(., \phi) \mid D)$ is
independent of the values $v({w}^i)$. As mentioned at the beginning of
this section, this is a typical feature of FB MC estimators.

This feature does not mean that the sampling distribution is
immaterial in FB MC however. Even though it does not arise in making
the estimate, as Eq.~\ref{eq:bayes_samp} shows, $v$ helps determine
what the expected loss will be. Indeed, in principle at least,
Eq.~\ref{eq:bayes_samp} can be used to guide the choice of the
sampling distribution for Bayesian FB MC.  It can even be used this
way dynamically, at a midpoint of the sampling process, when one
already has some samples of $U(., \phi)$. Such a procedure for using
Eq.~\ref{eq:bayes_samp} to set $v$ dynamically amounts to what is
called `active learning' in the PL literature~\citep[see][]{frse97,daka05}.

We now generalize the foregoing to the case of a non-quadratic loss
function $L$. The Bayesian estimator produces the estimate
\begin{eqnarray}
{\tilde{\mathscr{L}}}_D &\triangleq& {\mbox{argmin}}_{\rho \in {\mathbb{R}}}
[\int dU_c(., \phi) \;
P(U_c(., \phi) \mid D) L[{\tilde{\mathscr{L}}}_{D}, \;\; {\mathscr{L}}_{U_c}]]
\end{eqnarray}
Given that the factual oracle is $U(., \phi)$, the expected loss with that Bayesian
estimator is
\begin{eqnarray}
\int d{w}^1 \ldots d{w}^{N} \; \prod_{i=1}^{N} v({w}^i)
L[{\tilde{\mathscr{L}}}_{\{{w}^i, U({w}^i, \phi)\}}, \;\; \int d{w}' \; U({w}', \phi) ].
\label{eq:exp_loss}
\end{eqnarray}

The expected loss in Eq.~\ref{eq:exp_loss} is an average over data
with the oracle held fixed. This contrasts with the analogous quantity
typically considered in Bayesian analysis, which is an average over
oracles with the data held fixed. That quantity is the posterior
expected loss,
\begin{eqnarray}
\int dU(., \phi) P(U(., \phi) \mid D) L[{{\tilde{\mathscr{L}}}_D, \mathscr{L}}_U(\phi)]
\end{eqnarray}

In general, different $U(., \phi)$'s will give different risks for the
same estimator. So we can adapt any measure concerning loss in which
$U(., \phi)$ varies, to concern risk instead. In particular, the
posterior expected risk is
\begin{eqnarray}
\int dU(., \phi) P(U(., \phi) \mid D) \; \{L[{\tilde{\mathscr{L}}}_D,
{\mathscr{L}}_U(\phi)] \;-\; {\mbox{min}}_{\rho \in {\mathbb{R}}} 
[L[\rho, {\mathscr{L}}_U(\phi)]] \}.
\end{eqnarray}
Often the lower bound on loss is always 0, so that ${\mbox{min}}_{\rho
\in {\mathbb{R}}} [L[\rho, {\mathscr{L}}_U(\phi)]] = 0 \; \forall \;
U(., \phi)$. In this case posterior expected risk just equals
posterior expected loss.

We can combine the non-Bayesian and Bayesian analyses, involving
expected loss and posterior expected loss respectively. To do this we
consider the prior-averaged expected loss, given by
\begin{eqnarray}
\int dU(., \phi) \; P(U(., \phi)) \int d{w}^1 \ldots d{w}^{N} \;
\prod_{i=1}^{N} v({w}^i) L[{\tilde{\mathscr{L}}}_{\{({w}^i, U({w}^i,
\phi))\}}, \; \int d{w}' \; U({w}', \phi)].
\label{eq:ave_exp_loss}
\end{eqnarray}
where $P(U(., \phi))$ is a prior distribution over
oracles. 

Note that the prior-averaged expected loss is an average over both
oracles and sample sets.  It reflects the following experimental test
of our FB MCO algorithm: Multiple times a factual oracle $U(., \phi)$
is generated by sampling $P(U(., \phi))$. For each such $U(., \phi)$,
many times a factual sample set $D$ is generated by sampling the
likelihood $P(D
\mid U(., \phi), v)$. That $D$ is then used by the FMCO algorithm to
calculate ${\mathscr{L}}_D$. In performing that calculation, the
algorithm assumes the same likelihood as was used to generate $D$, but
its prior $P(U_c(., \phi))$ may not be the same function of $U_c(.,
\phi)$ as $P(U(., \phi))$ is of $U(., \phi)$. Then the loss between
${\mathscr{L}}_D$ and ${\mathscr{L}}_U$ is calculated.  The quantity
in Eq.~\ref{eq:ave_exp_loss} is the average of that loss.

Say that $P(U(., \phi))$ is the same function of $U(., \phi)$ as $P(U_c(.,
\phi))$ is of $U_c(., \phi)$. 
Then the Bayesian estimator is based on the actual prior. In this
case, the Bayesian estimator ${\tilde{\mathscr{L}}}_D$ will minimize
the prior-averaged expected loss of
Eq.~\ref{eq:ave_exp_loss}.{\footnote{To see this, replace
${\tilde{\mathscr{L}}}_D$ with some arbitrary function of $D$,
$f(D)$. Our task it to solve for the optimal $f$. First interchange
the integrals over data and over oracles in
Eq.~\ref{eq:ave_exp_loss}. Next consider the integrand of the outer
(data) integral,
\begin{eqnarray*}
\int dU(., \phi) \; P(U(., \phi)) 
\prod_{i=1}^{N} v({w}^i) L[f(D), \; \int d{w}' \; U({w}', \phi)] ].
\end{eqnarray*}
Since we are considering a noise-free oracle, we can write this as
\begin{eqnarray*}
\int dU(., \phi) \; P(U(., \phi)) P(D \mid v, U(., \phi))
L[f(D), {\mathscr{L}}_U(\phi)].
\end{eqnarray*}
Since $P(U(., \phi)) P(D \mid v, U(., \phi)) \propto P(U(., \phi) \mid
v, D) = P(U(., \phi) \mid D)$, this integral is minimized by setting
$f(D) = {\tilde{\mathscr{L}}}_D$. {\bf{QED.}}}}  In general though,
there is no reason to suppose that these two priors are the same. In
the real world where those priors differ, expected loss for a Bayesian
estimator is given by an inner product between the posterior used by
that estimator, $P(U_c(., \phi) \mid D)$, and the true posterior,
$P(U(., \phi) \mid D)$~\citep[see][]{wolp97,wolp96c}.{\footnote{It is in
recognition of the fact that those functions might differ that we have
been referring to `Bayesian' rather than `Bayes-optimal'
estimators.}}

As before, since $U(. \phi)$ varies in the integrand of
prior-averaged expected loss, we can can adapt it to get a
prior-averaged expected risk. This is given by
\begin{eqnarray}
&& \!\!\!\!\!\!\!\!\!\!\!\!\! \int dU(., \phi) \; P(U(., \phi)) \int d{w}^1 \ldots d{w}^{N} \;
\prod_{i=1}^{N} v({w}^i) 
\;\; \times \nonumber \\
&& \;\;\;\;\;\;\;\;\;\;\;\;\;\{L[{\tilde{\mathscr{L}}}_{\{({w}^i, U({w}^i,
\phi))\}}, \; {\mathscr{L}}(\phi)] \;\;-\;\; 
{\mbox{min}}_{\rho \in {\mathbb{R}}} [L[\rho, \;  {\mathscr{L}}(\phi)]] \}.
\end{eqnarray}
As before, if the minimal loss is always 0, then prior-averaged
expected risk just equals prior-averaged expected loss.

Broadly speaking, in Bayesian approaches to Monte Carlo problems, the
sampling distribution that generated the samples is immaterial once
one those samples have been generated ~\citep[see][]{ragh03} and
references therein). So what difference does the choice of a sampling
distribution like $v$ make to a Bayesian? The answer is that $v$
determines how likely it is that we will generate a $D$ with a high
posterior variance of the quantity of interest. For example, say one
wishes to form an importance sampling estimate of ${\mathscr{L}} =
\int dx \; U(x)$ using sampling distribution $v$ to generate sample
set $D$. Then if one changes $v$, one changes the likelihoods of the
possible $D$. Moreover, each $D$ has its own posterior variance,
Var$({\mathscr{L}}_c \mid D)$. So what a good choice of $v$ means is
that a $D$ with poor Var$({\mathscr{L}}_c
\mid D)$ is unlikely to be formed, that is, that $\int dD \; P(D \mid v)
{\mbox{Var}}({\mathscr{L}}_c \mid D)$ is low.

\subsection{Statistical Analysis of FB MCO}
\label{sec:exp_loss_mco}

We can extend the statistical analysis of FB MC to the case of FB MCO
by allowing $\phi$ to vary. The Bayesian choice of $\phi$ is the one
that minimizes posterior expected loss,
\begin{eqnarray}
{\tilde{\phi}}_D &\triangleq& {\mbox{argmin}}_\phi [\int dU_c \; P(U_c
\mid D) L(\phi, U_c)].
\end{eqnarray}
Since $P(U_c \mid v, D) = P(U_c \mid D)$, this estimator is
independent of $v$.  The same is true for the posterior expected loss
of this Bayesian estimator,
\begin{eqnarray}
\int dU \; P(U \mid D) L({\tilde{\phi}}_{D},  U).
\end{eqnarray}

On the other hand, the expected loss associated with this estimator,
\begin{eqnarray}
\int d{w}^1 \ldots d{w}^{N} \prod_{i=1}^{N} v({w}^i) L({\tilde{\phi}}_{\{{w}^i,
U({w}^i)\}},  U),
\end{eqnarray}
explicitly depends on $v$. So does the prior-averaged expected loss,
\begin{eqnarray}
\int dU \; P(U) \int d{w}^1 \ldots d{w}^{N} \prod_{i=1}^{N} v({w}^i) L({\tilde{\phi}}_{\{{w}^i,
U({w}^i)\}},  U).
\end{eqnarray}

Next, the posterior expected risk is
\begin{eqnarray}
\int dU \; P(U \mid D) \{L({\tilde{\phi}}_{D},  U) \;\;-\;\;
{\mbox{min}}_{\phi'} [L(\phi',  U)] \}
\end{eqnarray}
where $\phi'$ runs over the (implicit) set of all possible $\phi$.  In
general ${\mbox{min}}_{\phi'} [L(\phi', U)]$ varies with $U$. (For
example, this is the case with the loss function
${\mathscr{L}}_U(\phi)$ of Eq.~\ref{eq:mco_loss}.) Accordingly, unlike
in Bayesian FB MC, typically in Bayesian FB MCO the posterior expected
risk does not equal the posterior expected loss.

Finally, the prior-averaged expected risk is
\begin{eqnarray}
\int dU d{w}^1 \ldots d{w}^{N} \;  P(U) \prod_{i=1}^{N} v({w}^i)
\{L({\tilde{\phi}}_{\{{w}^i, U({w}^i)\}},  U) \;\;-\;\;
{\mbox{min}}_{\phi'} [L(\phi',  U)] \}.
\label{eq:pr_ave_exp_risk}
\end{eqnarray}
Again, since ${\mbox{min}}_{\phi'} [L(\phi', U)]$ typically varies
with $U$, in general this prior-averaged expected risk does not equal
the prior-averaged expected loss. However the estimator that minimizes
prior-averaged expected loss --- ${\tilde{\phi}}_D$ --- is the same as
the estimator that minimizes prior-averaged expected
risk.{\footnote{This follows from the fact that the prior-averaged
lowest possible risk, the term subtracted in
Eq.~\ref{eq:pr_ave_exp_risk}, is independent of the choice of the
estimator.}}

For any particular fitting algorithm, our equations tell us how
performance of the associated FB MCO depends on $v$ and either
$P(U(. ,\phi))$ or the pair $P(U(., \phi))$ and $P(U_c(.,
\phi))$, depending on which equation we consider. So if we fix those
prior(s), our equations tell us, formally, what the optimal $v$
is. 

One can consider estimating that optimal $v$ at a mid-way point of the
algorithm, based on the algorithm's behavior up to that point. One can
then set $v$ to that estimate for the remainder of the
algorithm.{\footnote{Note though that if one intends to update $v$
more than once, then strictly speaking the first update to $v$ should
take into account the fact that the future update will occur. That
means the equations above for expected loss, prior-averaged expected
loss, etc., no longer apply.}} Doing this essentially amounts to
a type of active learning.

As with Bayesian FBMC, we can analyze the effects of having $P(U)$ not
be the same function of $U$ as $P(U_c)$ is of $U_c$.  Since PL and MCO
are formally the same, such an analysis applies to parametric machine
learning in addition to FB MCO. In particular, the analysis gives a
Bayesian correction to the bias-variance decomposition of supervised
learning. This correction holds even if the fitting algorithm in the
supervised learning cannot be cast as Bayes-optimal for some assumed
prior $P(U_c)$. Intuitively speaking, the correction means that the
bias-variance decomposition gets replaced by a
bias-variance-covariance decomposition. That covariance is between the
posterior distribution over target functions on the one hand, and the
posterior distribution over fits produced by the fitting algorithm on
the other~\citep[see][]{wolp97}.

\subsection{Combining FB and Non-FB Estimates in FB MCO} 

Return now to the example in Sec.~\ref{sec:exp_loss_mco}, where the
factual sample is formed by importance sampling the factual oracle and
we form a fictitious sample set using fictitious oracles.Then using
only ${D}$, our estimate of ${\mathscr{F}}_G(\theta)$ would be the
factual estimate, ${\hat{\mathscr{F}}}_{{D}}(\theta) =
\frac{\sum_{k=1}^N r^k(\theta)}{N}$.  Using only our fictitious
samples would instead give us the estimate
${\hat{\tilde{\mathscr{F}}}}_D(\theta)$.

On the one extreme, say we firmly believe that distribution we use for
the posterior $P({G_c} \mid D)$ is correct. (So in particular we
firmly believe that the factual oracle $G$ was generated by sampling
the prior $P(G_c)$.) Then in the limit $N_T
\rightarrow \infty$, $\forall \theta$ our importance-sample estimate
of ${\tilde{{\mathscr{F}}}}_{D}$ will be exactly correct. So Bayesian
decision theory would direct us to use the associated estimate
${\hat{\tilde{\mathscr{F}}}}_D(\theta)$, and ignore
${\hat{\mathscr{F}}}_{{D}}(\theta)$. At the other extreme, say that
$N_T = 1$, while $N$, the number of factual samples, is quite
large. In such a situation, even if we believe our posterior is
correct, it would clearly be wrong to use
${\hat{\tilde{\mathscr{F}}}}_{D}(\theta)$ as our estimate, ignoring
${\hat{\mathscr{F}}}_{{D}}(\theta)$.

How should we combine the estimates in this latter situation?  More
generally, even when we believe our posterior is correct, unless the
number of fictitious samples is far greater than the number of factual
samples, we should combine the two associated estimates. How best to
do that? Does the fact that ${\hat{\tilde{\mathscr{F}}}}_D(\theta)$ is
estimated via importance sampling over a much larger space than
${\hat{\mathscr{F}}}_{{D}}(\theta)$ affect how we should combine them?
More generally, say we don't presume that our $P({G_c}
\mid {{D}})$ is exactly correct; how should we combine the
estimates then?

One is tempted to invoke Bayesian reasoning to determine how best to
combine the two estimates. While that might be possible in certain
situations, often determining the optimal Bayesian combination would
necessitate yet more Monte Carlo sampling of some new integrals. It
would be nice if some other approach could be used.

One potential such approach is
stacking~\citep{wolp92,brei96,smwo99}. In this approach, one many times
partitions the factual sample $D$ into two parts, a `training set'
$D^1$, and a `validation set' $D^2$. We write the values of $w$ and
$U$ in $D^1$ as $\{D_w^1(i)\}$ and $\{D_U^1(i, \phi)\}$ respectively,
and similarly for $D^2$. For each such partition one would run both
the non-FB MCO algorithm and the FB MCO algorithm on $D^1$. That
generates the estimates ${\hat{\phi}}_{v,U,D^1}$ and
${\hat{\tilde{\phi}}}_{D^1}$, respectively.

Those two $\phi$'s give us two associated error values on the
validation set, $\sum_j D_U^2(j, {\hat{\phi}}_{v,U,D^1})$ and $\sum_j
D_U^2(j, {\hat{\tilde{\phi}}}_{D^1})$, respectively.  More generally,
we can evaluate the error on the the validation set of $any$ $\phi$,
in addition to the errors of ${\hat{\phi}}_{v,U,D^1}$ and
${\hat{\tilde{\phi}}}_{D^1}$. Moreover, we can do this for the
validation set of any of the partitions of $D$. Note, however, that \emph{only} factual samples are used for cross-validation.

This is what stacking exploits.  In the most straightforward use of
stacking, one searches for a function mapping the $\phi$'s produced by
our two algorithms to a composite $\phi$. The goal is to find such a
composite $\phi$ that will have as small validation set error (when
averaged over all partitions) as possible.

For example, if $\phi$ is a Euclidean vector, one could perform a
regularized search for the weighted sum of $\phi$'s that gives minimal
partition-averaged validation set error. Let the weights produced by
that search be $b_{FB}$ and $b_{non-FB}$. Then to find the final
estimate for $\phi$, one would use those weights to sum the outputs of
the algorithms when run on all of $D$: $b_{non-FB}
{\hat{\phi}}_{v,U,D} \;+\; b_{FB}{\tilde{\phi}}_{D}$.

\section{Conclusion}

In this paper we explored the relationship between Monte Carlo
Optimization of a parametrized integral, parametric machine
learning, and `blackbox' or `oracle'-based optimization. We made
four contributions. 

First, we proved that MCO is identical to a broad class of parametric
machine learning problems. This should open a new application domain
for previously investigated parametric machine learning techniques, to
the problem of MCO.  

To test the use of PL in MCO one needs an MCO problem domain. The one
we used was based on our second contribution, which was the
introduction of immediate sampling.  Immediate sampling is a way to
transform an arbitrary blackbox optimization problem into an MCO
problem.  Accordingly, it provides us a way to test the use of PL to
improve MCO, but testing whether it can improve blackbox optimization.

In our third contribution we validated this way of improving blackbox
optimization. In particular, we demonstratied that cross-validation
and bagging improve immediate sampling. 

Conventional Monte Carlo and MCO procedures ignore some features of
the sample data. In particular, they ignore the relationship between
the sample point locations and the associated values of the integrand;
only the values of the integrand at those locations are considered. We
ended by presenting fit-based MCO, which is a way to exploit the
information in the sample locations.

There are many PL techniques that should be applicable to immediate
sampling but that are not experimentally tested in this paper. These
include density estimation active learning, stacking, kernel-based
methods, boosting, etc. Current and future work involves experimental
tests of the ability of such techniques to improve MCO in general and
immediate sampling in particular.

Other future work is to conduct experimental investigations of the
three techniques that we presented in this paper but did not test. One
of these is fit-based MCO (and fit-based immediate sampling in
particular). The other two are the techniques described in the
appendices: immediate sampling for constrained optimization problems,
and immediate sampling with elite objective functions.

There are also many potential application domains for immediate
sampling PC for blackbox optimization that we intend to explore. Some
of these exploit the ability of such PC to handle arbitrary (mixed)
data types of $x$'s. In particular, one such data type is the full
trajectory of a system through a space; for optimizing a problem over
such a space, PC becomes a form of reinforcement learning.

\begin{appendices}
\section{Constrained Optimization}

Under the PC transform we replace an optimization problem over $X$
with one over ${\mathscr{Q}}$. As discussed at the beginning of Sec.~\ref{subsec:AdvantagesOfPC}, the characteristics of the transformed objective can be very different from those of the original objective.

Similarly, characteristics of any constraints on $X$ in the
original problem can also change significantly under this transformation. More
precisely, say we add to (P4) equality and inequality constraints restricting $x \in X$ to a {\bf{feasible region}}. Then to satisfy  those $X$-constraints we need to modify (P5) to ensure that the
support of the solution $q_\theta(.)$ is a subset of the feasible
region in $X$.

This appendix considers some ways of modifying PC to do this. For
earlier work on this topic in the context of delayed sampling, see~\citet{wost06,biwo04a,biwo04b,mawo05}.

%

\subsection{Guaranteeing Constraints}

Say we have a set of equality and inequality constraints over $X$.
Indicate the feasible region by a feasibility indicator function 
\begin{equation*}
\Phi(x) = \left\{
\begin{array}{ll}
1, & x \textrm{ is feasible,}\\
0, & \textrm{otherwise.}
\end{array}
\right.
\end{equation*}
For simplicity, we assume that for any $x$, we can evaluate $\Phi(x)$ essentially `for free'. 

The transformed version of this constrained optimization problem is
\begin{equation*}
\begin{array}{lll}
(\textrm{P5}_c):& \textrm{minimize}&\mathscr{F}_{\mathscr{G}}(q_{\theta}),\\
&\textrm{subject to}&q_{\theta}(x)\Phi(x) = 0.
\end{array}
\end{equation*}
We now present a parametrization for $q$ that ensures that it has zero
support over infeasible $x$.  First, let ${\tilde{q}}$ be any
parametrized distribution over $X$, for instance, a mixture of
Gaussians. Then using $\Phi(x \in X)$ as a `masking funtion' we
parametrize $q_\theta(x)$ as
\begin{eqnarray*}
q_\theta(x) &\triangleq& \frac{{\tilde{q}}_\theta(x)\Phi(x)}{\int dx' \;
{\tilde{q}}_\theta(x') \Phi(x')}  \nonumber \\
&\triangleq& {\tilde{q}}_{\Phi, \theta}(x).
\end{eqnarray*}
This $q_\theta$ automatically meets the constraints; it places
zero probability mass at infeasible $x$'s.  It transforms the constrained problem (P5)$_c$ into the unconstrained problem 
\begin{equation*}
(\textrm{P5}_{uc}): \textrm{minimize }
{\mathscr{F}}_{{\mathscr{G}}}({\tilde{q}}_{\Phi,\theta}).
\end{equation*}

Now consider the case where ${\mathscr{F}}_{{\mathscr{G}}}$ is an
integral over $X$. Typically in this case we are only concerned with
the values of the associated integrand at feasible $x$'s. For example,
when $E_{q_\theta}(G)$ is of interest, it's usually because our
ultimate goal is to find a feasible $x$ with as good a $G(x)$ as
possible. In this situation it makes no sense to choose between two
candidate $q_\theta$'s based on differences in (the $G$ values at) the
regions of infeasible $x$ that they emphasize. More formally, our
choice between them should be independent of their respective values
of $\int dx \; [1 - \Phi(x)] q_\theta(x) G(x)$.  We can enforce this
by replacing the objective $E_{q_\theta}(G) = \int dx \; q_\theta(x)
G(x)$ with $\int dx \;
\Phi(x) q_\theta(x) G(x)$. If we then use the barrier function
approach outlined above, our final objective becomes $qp$ KL distance
with the integral restricted to feasible $x$'s.

Generalizing this, when we are not interested in behavior at
infeasible $x$ we can reduce the optimization problem further from
(P5$_{uc}$), by restricting the integral to only run over feasible
$x$'s.  More precisely, write the original problem (P5$_c$) as
the minimization of
\begin{eqnarray*}
\int dx [\int dg P(g \mid x, {\mathscr{G}}) ] F(g, q_\theta(x))
&\triangleq& \int dx \; \mu(x, q_\theta(x)) ,
\end{eqnarray*}
subject to the constraints on the support of $q_\theta$.  By using the
${\tilde{q}}$ construction we can replace this constrained
optimization problem with the unconstrained problem
\begin{eqnarray*}
\textrm{(A1): }{\mbox{argmin}}_{q} \; \int dx \; \Phi(x) \mu[x, q_\theta(x)]
&=&
{\mbox{argmin}}_{{\theta}} \int dx \;\Phi(x) \mu[x, {\tilde{q}}_{\Phi,\theta}(x)],\\
&=& {\mbox{argmin}}_{{\theta}} \int dx \;\Phi(x) \mu[x, \frac{\Phi(x){\tilde{q}}(x)}{\int dx' \Phi(x')
{\tilde{q}}(x')}].
\end{eqnarray*}

As an example, say our original objective function is $pq$ KL
distance. Define $Z^\beta_\Phi \equiv \int dx \; p^\beta(x)
\Phi(x)$. Then our new optimization problem is to minimize over $\theta$
\begin{eqnarray*}
{\mbox{KL}}(p^\beta_\Phi \; || \; {\tilde{q}}_{\Phi, \theta}) &=&
{\mbox{KL}}(\frac{p^\beta \Phi}{Z^\beta_\Phi} \; || \;
{\tilde{q}}_\Phi) \nonumber \\ &=& -\int dx \; \frac{\Phi(x)
p^\beta(x)}{Z^\beta_\Phi}
{\mbox{ln}}[\frac{{{\tilde{q}}_\theta}(x)\Phi(x)}{\int dx'
{{\tilde{q}}_\theta}(x')
\Phi(x')}],\nonumber\\
&=& -\int dx \; \frac{\Phi(x) p^\beta(x)}{Z^\beta_\Phi} \; \{ {\mbox{ln}}[{{\tilde{q}}_\theta}(x)] +
{\mbox{ln}}[\Phi(x)] - {\mbox{ln}}[\int dx' {{\tilde{q}}_\theta}(x') \Phi(x')]\}.
\end{eqnarray*}
The ${{\tilde{q}}_\theta}$ minimizing this is the same as the one that maximizes
\begin{eqnarray}
 \int dx \frac{\Phi(x) p^\beta(x)}{Z^\beta_\Phi}  {\mbox{ln}}[{{\tilde{q}}_\theta}(x)] &-&
{\mbox{ln}}[\int dx' \; {{\tilde{q}}_\theta}(x') \Phi(x')].
\label{eq:ensure_feas}
\end{eqnarray}
We can estimate $Z^\beta_\Phi$ using MC techniques. We can then apply
MCO to estimate the $\theta$ that maximizes the integral difference{\footnote{Note that in general this
difference of integrals will not be convex in ${{\tilde{q}}_\theta}$
for product distributions, unlike KL$(p^\beta_\Phi \; || \;
{{\tilde{q}}_\theta})$. See the discussion at the end of
Sec.~\ref{sec:overview} on product distributions and $pq$ distance.}}
 in Eq.~\ref{eq:ensure_feas}.

To generate a sample of sample $q_\theta(x) =
{{\tilde{q}}_\theta}(x)\Phi(x)$ we can subsample{\footnote{Say we want to sample a distribution
$A(x) \propto B(x) C(x)$ where $B$ is a distribution and $C$ is
non-negative definite, with $c$ some upper bound on $C$. To generate
such a sample by `subsampling $B$ according to $C$' we first
generate a random sample of $B(.)$, getting $x'$. We then toss a coin
with bias $C(x') / c$. If that coin comes up heads, we keep $x'$ as
our sample of $A$. Otherwise we repeat the
process~\citep[see][]{wost06,roca04}.}}  ${\tilde{q}}_\theta$
according to $\Phi$. In some cases though, 
this can be very inefficient (that is, one
may get many rejections before getting a feasible $x$). To deal with
such cases, we can first run a density estimator on the samples of
feasible $x$'s we have so far, getting a distribution $\pi$. (Note
that no extra calls to the feasibility oracle are needed to do this.)
Next write $q_\theta(x) = \pi(x) [q_\theta(x) /
\pi(x)]$. This identity justifies the generation of samples of
$q_\theta$ by first sampling $\pi(x)$ and then subsampling according
to $q_\theta(x) / \pi(x) = {\tilde{q}}_\theta(x) \Phi(x) / \pi(x)$.

In an obvious modification to the foregoing, we can replace the hard
restriction that supp($q$) contain only feasible $x$'s, with a `soft'
constraint that $q(x) \le \kappa \; \forall \; {\mbox{infeasible}} \;
x$. A similar alternative is to `soften' $\Phi(x)$ by replacing it
with $\kappa$ for all infeasible $x$, for some $\kappa > 0$.  For
either alternative we anneal $\kappa$ down to 0, as usual, perhaps
using cross-validation.

\subsection{Alternative ${\mathscr{F}}_{{\mathscr{G}}}$}

Since we're maximizing our expression over ${{\tilde{q}}_\theta}$, the
second, correcting integral in Eq.~\ref{eq:ensure_feas} will tend to
push ${{\tilde{q}}_\theta}$ to have probability mass $away$ from
feasible regions. To understand this intuitively, say that
${{\tilde{q}}_\theta}$ is a Gaussian and that the feasible region is
`spiky', resembling a multi-dimensional star-fish with a large central
region and long, thin legs. For this situation, if we
over-concentrate on keeping most of ${{\tilde{q}}_\theta}$'s mass
restricted to feasible $x$, our Gaussian will be pushed away from any
of the spikes of the feasible $x$'s, and concentrate on the center. If
the solution to our original optimization problem is in one of those
spikes, such over-concentration is a fatal flaw. The second integral
in Eq.~\ref{eq:ensure_feas} corrects for this potential problem.

More broadly, consider typical case behavior when one applies some
particular constrained optimization algorithm to any of the problems
in a particular class of optimization problems.  As a practical
matter, there is a spectrum of such problem classes, indexed by how
difficult it is just to find feasible solutions on typical problems of
the class.  On the one side of this spectrum are problem classes where
it is exceedingly difficult to find such a solution, e.g.,
high-dimensional satisfiability problems with a performance measure
$G$ superimposed to compare potential solutions.  On the other end are
``simple'' problem classes where it is reasonable to expect to find a
feasible solution.  The `starfish' optimization problem is an example
of a problem of the former type.{\footnote{Note that since we are
discussing typical-case behavior, computational complexity
considerations do not apply.}}

For problems on the first side of the spectrum, where just getting a
substantial amount of probability mass into the feasible region is
very difficult, we may want to leave out the second integral in
Eq.~\ref{eq:ensure_feas}. In other words, we may want to minimize
KL$(p^\beta_\Phi \; || \; {\tilde{q}}_\theta)$ rather than
KL$(p^\beta_\Phi \; || \; {\tilde{q}}_{\Phi,\theta})$.  The reason to
make this change is so that ${\tilde{q}}_\theta$ won't get pushed away
from the feasible region. (As an aside, another potential benefit of
this change is that if we make it, then for product distribution
${\tilde{q}}_\theta$, ${\mathscr{F}}_{{\mathscr{G}}}(.)$ is convex.)

Even if we do make this change, when we sample the resultant
${\tilde{q}}_\theta$ we may not get a feasible $x$. If this happens, a
natural approach is to repeatedly sample ${\tilde{q}}_\theta$ until we
do get a feasible $x$.  However the resultant distribution of $x$'s is
the same as that formed by sampling ${\tilde{q}}_{\Phi,\theta}$ for
the same $\theta$. So under this 'natural approach' we work to
optimize a distribution (${\tilde{q}}_\theta$) different from the one
we ultimately sample (${\tilde{q}}_{\Phi,\theta}$).  This means that
this approach may not properly balance our two conflicting needs for
${\tilde{q}}_\theta$: that it have most of its support in the feasible
region, and that it be peaked about $x$'s with high $p^\beta(x)$.

To illustrate this issue differently, take ${\tilde{q}}_{\theta}$ to
be normalized, and to avoid multiplying and dividing by zero, modify
$\Phi(x)$ to equal some very small non-zero value $\kappa$ for
infeasible $x$ (as discussed above). Then under this `compound
procedure', we ultimately sample
${\tilde{q}}_{\Phi,\theta}(.)$. However we do not choose
${\mathscr{F}}_{{\mathscr{G}}}({\tilde{q}}_{\Phi,\theta}) =$
KL$(p^\beta_\Phi \; || \; {\tilde{q}}_{\Phi,\theta})$ as the function
of $\theta$ that we want to minimize. Instead we
choose
\begin{eqnarray}
{\mathscr{F}}_{{\mathscr{G}}}( {\tilde{q}}_{\Phi,\theta}) &=&
{\mbox{KL}}(p^\beta_\Phi \; || \; 
 {\tilde{q}}_{\Phi,\theta}) \;-\; {\mbox{ln}}[\int dx' \; \frac{ {\tilde{q}}_{\Phi,\theta}(x')}{
\Phi(x')}] .
\end{eqnarray}

\subsection{Using Constraints for Unconstrained Optimization}

Return now to unconstrained optimization problems. Say that we have
reason to expect that over a particular region $R$, the distribution
$p^\beta(x)$ has values approximately $\kappa$ times as small as its
value over $X \setminus R$. It would be nice to reflect this insight
in our parametrization of $q$, that is, to parametrize $q$ in a way that
makes it easy to match it to $p^\beta(x)$ accurately.  We can do this
using a binary-valued function $\Phi$ and the approaches presented above.

To illustrate this, define ${\tilde{q}}_{\Phi,\theta}$ as above and
choose the objective function ${\mathscr{F}}_{\mathscr{G}}(\theta) =$
KL$(p^\beta \;||\; {\tilde{q}}_{\Phi,\theta})$, where $\Phi(x) =
\kappa$ over $R$, and equals 1 over $X \setminus R$.{\footnote{Note that we use
$p^\beta$ in this ${\mathscr{F}}_{\mathscr{G}}$, not $p^\beta_\Phi$,
which is what we used for constrained optimization. This is because
our goal now is simply to find a $q_\theta$ that matches
$p^\beta(x)$. There are no additional aspects to the problem involving
feasibility regions that have no {\it{a priori}} relation to $G(x)$.}}
Then working through the algebra, the $q_\theta$ that minimizes this
objective is given by the ${{\tilde{q}}_\theta}$ that minimizes
\begin{eqnarray}
-\int dx p^\beta(x)  {\mbox{ln}}[{{\tilde{q}}_\theta}(x)] \;+\; {\mbox{ln}}[\int dx'
{{\tilde{q}}_\theta}(x')
\Phi(x')] &=& {\mbox{KL}}(p^\beta \;||\; {{\tilde{q}}_\theta}) \;+\; {\mbox{ln}}[\int dx' {{\tilde{q}}_\theta}(x')
\Phi(x')]  \nonumber \\
&=& {\mbox{KL}}(p^\beta \;||\; {{\tilde{q}}_\theta}) \;+\ \nonumber \\
&& \;\;\;\;\;{\mbox{ln}}[\int dx'_R \; \kappa {{\tilde{q}}_\theta}(x') +  \int dx'_{X
\setminus R} \; {{\tilde{q}}_\theta}(x')] . \nonumber \\
&&
\end{eqnarray}
The logarithm on the right-hand side is a `correction' to $pq$
distance from $p^\beta$ to ${{\tilde{q}}_\theta}$, a correction that
pushes ${{\tilde{q}}_\theta}$ $away$ from regions where $\Phi(x) = 1$
(assuming $\kappa < 1$). To use immediate sampling with this
parametrization scheme, once we find the ${{\tilde{q}}_\theta}$ that
minimizes the sum of $pq$ distance plus that correction term, we would
set $h$ to the distribution (proportional to)
${{\tilde{q}}_\theta}(x)\Phi(x)$. So we would generate our new samples
from ${{\tilde{q}}_\theta}(x)\Phi(x)$, for example by
subsampling.{\footnote{In practice, $\Phi(x)$ for this unconstrained
case would not be provided by an oracle. Instead we would typically
have to estimate it. We could do that for example by using a
regression to form a fit to samples of $p^\beta(x)$ and then use that
regression to define the region $R$.}}

\section{The Elite Objective Function}

Not all PC objectives can be cast as an integral transform. Properly
speaking, the choice of objective should be set by how the final
$q_\theta$ will be used. For instance, the concept of expected improvement suggested by ~\citet{moti78}, and used by \citet{josc98}, considers an objective (to be maximized) given by $\max(G_{\mathrm{bc}} - G(x), 0)$, where $G_{\mathrm{bc}}$ is the best of all the current samples, $\min_i \{G(x^i)\}$. This means that at each step we will take a single sample, and want to maximize the improvement. This is a simplification; even though the next sample may yield any improvement, it may be informative, so that we get a good sample ten steps later. A less simplistic objective is the following: 

In blackbox optimization, no matter how many calls to the (factual) oracle we make, we will ultimately choose the best $x$ (as far as the associated $G$  value is concerned) out of all the ones that were fed to the oracle during the course of the entire run. Our true goal in BO is to have the $G$ associated with $that$ best $x$ be as small as possible. For a discussion of distributions of extremal values, see ~\citet[][]{leli83,resn87}. 
 
Given that $q_\theta$ varies over the run in a way that we do not 
know beforehand, how can one approximate this goal as minimizing an 
objective function that is well-defined at all points during the run? 
One way to do this is to assume that there is some integer $N$ such 
that, simultaneously, 
\begin{enumerate}  
\item It is likely that the best $x$ will be one of the final $N$ 
calls to the oracle during the run; 
\item It is likely that $q_\theta$ will not vary much during the 
generation of those final $N$ samples. 
\end{enumerate} 
Under (2) we can approximate the $q_\theta$'s that are used to 
generate the final $N$ calls as all being equal to some canonical 
$q_\theta$. Under (1), our goal then becomes finding the canonical 
$q_\theta$ that, when sampled $N$ times, produces a set of $x$'s whose 
best element is as good as possible. {\footnote{An obvious variant of this reasoning is to have 
$N$ vary across the run of the entire algorithm, at any iteration $t$ 
being only the number of {\it{remaining}} calls to the oracle that we 
presume will be made. In this variant, one would modify the elite 
objective function to only involve the $N(t)$  remaining samples whose 
$G$ value is better than the best found by iteration $t$. For the case $N=1$, this is 
analogous to the expected improvement idea in~\citet{josc98}. Note 
that this variant objective function will change during the run, which 
may cause stability problems.}}

In this appendix we make some cursory comments about this objective
function, which we call the {\bf{elite objective function}}. We focus
on the use of Bayesian FB techniques with this objective. For a noise-free oracle the CDF for the elite objective is
\begin{eqnarray}
{\mbox{CDF}}(k) &\triangleq& 1 - \int dx^1 \ldots dx^N \; \prod_{i=1}^N
[q_\theta(x^i) \Theta(G(x^i) - k)].
\end{eqnarray}
So the associated density function is
\begin{eqnarray}
f(k) &=& \frac{d \; {\mbox{CDF}}(k)}{dk} \nonumber \\
&=& N q_\theta(k) [\int dx \; q_\theta(x) \Theta(G(x) - k)]^{(N-1)}.
\end{eqnarray}
The associated expectation value, $\int dk \; k f(k)$, is not linear
in $q_\theta$.

%
%


Writing it out, the posterior expected best-of-$K$ value returned by
the oracle when queries are generated by sampling $q_\theta$ is
\begin{eqnarray}
\int dx^1 \ldots dx^K \; \prod_{i=1}^K q_\theta(x^i) \int dG \;
P(G \mid D) \; \int dg^1 \ldots dg^K \prod_{j=1}^K P(g^k \mid x^k, G)
{\mbox{min}}_k \{g^k\}
\label{eq:elite}
\end{eqnarray}
We want the $\theta$ minimizing this. Say we knew the exact posterior
$P(G \mid D)$ and could evaluate the associated integral in
Eq.~\ref{eq:elite} closed-form. In this case there would be need for
the parametric machine learning techniques used in the text. In particular
there would be no need for regularization --- an analogous role is
played by the prior $P(G)$ underlying $P(G \mid D)$.

When we cannot evaluate the integral in closed form we must
approximate it.  To illustrate this, as in Sec.~\ref{sec:fit_based},
for simplicity consider a single-valued oracle $G$. This reduces
Eq.~\ref{eq:elite} to
\begin{eqnarray} 
\int dx^1 \ldots dx^K \; \prod_{i=1}^K q_\theta(x^i) \int dG \; 
P(G \mid D) {\mbox{min}}_k \{G(x^k)\}. 
\label{eq:elite_single_valued} 
\end{eqnarray} 
(The analogous FB MCO equation for objective functions involving a
single integral Eq.~\ref{eq:bayes_imm_single_valued}; here the single
`$x$' in that equation is replaced with a set of $K$ $x$'s sampled
from $q_{\theta}$.)  To approximate this integral we draw $N_T$
sample-vectors of $K$ $x$'s each, using a sampling distribution
$h_c(x)$ to do so. At the same time we draw $N_T$ fictitious oracles
from some sampling distribution $H$ over oracles.

To simplify notation, let ${\vec{x}}$ indicate such a $K$-tuple of
$x$'s. (So for multidimensional $X$, ${\vec{x}}$ is actually a
matrix.) Also write
\begin{eqnarray}
h_c({\vec{x}}) &\triangleq& \prod_{i=1}^K {h_c}(x^k), \nonumber \\
q_\theta({\vec{x}}) &\triangleq& \prod_{i=1}^K q_\theta(x^k), \nonumber \\
G({\vec{x}}) &\triangleq& (G(x^1), \ldots, G(x^K)). 
\end{eqnarray}
With this notation, the estimate based on fictitious samples
introduced in Sec.~\ref{sec:fit_based} becomes\footnote{In practice
there might be more efficient sampling procedures than
Eq.~\ref{eq:elite_sum}. For example, one could form $NK$ samples of
${h_c}(x)$ and $N$ samples of $H(G)$, to get two sets, which one then
subsamples many times, to get pairs $[{\vec{x}}, G({\vec{x}})]$.}
\begin{eqnarray}
\sum_{i = 1}^N
	\prod_{j=i}^K \frac{q_\theta(x^j_i)}{{h_c}(x^j_i)} \frac{P(G_i \mid D)}{H(G_i)}
{\mbox{min}}_{k=1,\ldots K} \{G_i(x^k_i)\} .
\label{eq:elite_sum}
\end{eqnarray}

As discussed in Sec.~\ref{sec:fit_based}, it is often good to set
$H(G)$ to be as close to $P(G \mid D)$ as possible. So for example if
we assume a Gaussian process model, typically we can set $H(G_i) =
P(G_i \mid D)$, and then directly sample $H$ to get the values of one
$G_i$ at the $K$ separate points $x^j_i$. Alternatively, we can first
form a fit $\phi(x)$ to the data in $D$. Next, for each of $N$ samples
${\vec{x}}_i$, sample a colored (correlated) noise process over the
$K$ points \{${\vec{x}}_i$\} to get $K$ real numbers.
Finally, add those $K$ numbers to the corresponding values
$\{\phi({\vec{x}}_i^j) : j = 1, \ldots, K\}$. This gives our desired
sample of \{$G_i({\vec{x}}_i)$\}.


To illustrate the foregoing, suppose $K = 1$, and that we have no
regularization on $q_\theta$. Then, in general, the sum in
Eq.~\ref{eq:elite_sum} is minimized by a $q_\theta$ that is a delta
function about that data point $x^1_i$ with the best associated value
$G_i(x^1_i) / {h_c}(x^1_i)$. However for $K > 1$, even without
regularization, the optimal $q_\theta$ is \emph{not} a delta function,
in general.\footnote{This suboptimality of a delta function $q_\theta$
is similar to the suboptimality of having all $K$ pulls in a
multi-armed bandit problem be pulls of the same arm.} In addition to the regularization-based argument in the text, this gives a
more formal reason why the optimal $q_\theta$ should not be infinitely
peaked.

%
%

When $K > 1$, the peakedness of $q_\theta$ parallels the peakedness of
another non-negative function over $x$'s, namely $P(G : G(x) {\mbox{
is minimized at }} x \mid D)$. However, if we run a few iterations of
FB MCO with the elite objective, then $D$ grows, and so $P(G : G(x)
{\mbox{ is minimized at }} x \mid D)$ gets increasingly peaked over
$x$'s. (Intuitively, the larger $D$ is, the more confident we are
about $G$, and consequently the more confident we are about what
regions of $x$'s minimize $G$.)  Accordingly, $q_\theta$ gets
increasingly peaked as the algorithm progresses. 

Note that this happens even though there is no external annealing
schedule. This reflects the fact that the elite objective has no
hyperparameter or regularization parameter like the $\beta$ that
appears in both the $pq$ and $qp$ objective functions.


\section{Gaussian Example for Risk Analysis}
\label{sec:GaussExample}

The following example illustrates the foregoing for the case of
Gaussian $\pi$, where only moments of $\pi$ up to order 2 matter.

To illustrate the foregoing, consider the
simple case where there are only two $\phi$'s, $\phi^1$ and $\phi^2$.
Suppose that $U$ and $X$ are such that $\pi$ is a two-dimensional
Gaussian. Write $\pi$'s mean as $\mu$. Say that one of $\pi$'s
principal axes is parallel to the diagonal line, $l_1 =
l_2$ (that is, one of the eigenvectors of $\pi$'s covariance
matrix is parallel to the diagonal, and one is orthogonal to the
diagonal). Write the standard deviation of $\pi$ along that diagonal
axis as $\sigma_A$, and write the standard deviation along the other,
orthogonal axis as $\sigma_B$.

Since $\pi$'s covariance matrix has identical diagonal entries, and
since the trace of that matrix is preserved under rotations, those
entries are both $\frac{1}{2}[\sigma_A^{2} + \sigma_B^{2}]$. Since the
determinant is preserved, and since $\sigma_A$ is the variance
parallel to the diagonal, this in turn means that $\pi$'s (identical)
off-diagonal entries are $\frac{1}{2}[\sigma_A^{2} -
\sigma_B^{2}]$. The probability that MCO will choose $\phi^1$ is the
integral of $\pi$ over the half-plane where $\phi^1 \le \phi^2$:
\begin{eqnarray}
{\mbox{Pr}}({{\hat{\mathscr{L}}}}(\phi^2) > {{\hat{\mathscr{L}}}}(\phi^1) ) &=&
\erf(\frac{\mu_2 - \mu_1}{\sigma_B \sqrt{2}}) .
\end{eqnarray}

Next, define
\begin{eqnarray}
\Delta L &\equiv&  {\mathscr{L}}(\phi^1) - {\mathscr{L}}(\phi^2), \nonumber \\
\Delta b &\equiv& [\mu_1 - {\mathscr{L}}(\phi^1)] \;\;-\;\; [\mu_2 -
{\mathscr{L}}(\phi^2)] \nonumber \\ 
	&=& [\mu_1 - \mu_2] - \Delta L .
\end{eqnarray}
So the difference in the value of the loss function between the two
$\phi$'s is $\Delta L$, and the the difference in the biases of the
two estimators ${{\hat{\mathscr{L}}}}(\phi^1)$ and ${{\hat{\mathscr{L}}}}(\phi^2)$ is $\Delta
b$.  Note also that the variances of the two estimators are the same,
\begin{eqnarray}
{\mbox{Var}}[{\mathscr{L}}(\phi^1)] \;\;=\;\; {\mbox{Var}}[{\mathscr{L}}(\phi^2)] \;\;=\;\; 
\frac{\sigma_A^2 + \sigma_B^2}{2}.
\end{eqnarray}
So if we shrink the variance of either of the estimators, then we
shrink an upper bound on $\sigma_B$. 

For this case of a fixed set of $\phi$'s, it is illuminating to
consider the difference between expected loss under a particular MCO
algorithm and minimal expected loss over all $\phi$'s, that is, the risk
of the MCO algorithm. Assuming $\Delta L < 0$, it is given by
\begin{eqnarray}
[{\mbox{Pr}}[{{\hat{\mathscr{L}}}}(\phi^2) > {{\hat{\mathscr{L}}}}(\phi^1 )]\;-\;
\Theta[{\mathscr{L}}(\phi^2) - {\mathscr{L}}(\phi^1)] ] \;\times \;
[{\mathscr{L}}(\phi^1) \;-\; {\mathscr{L}}(\phi^2)] \nonumber
\end{eqnarray}
\begin{eqnarray}
&=& \nonumber
\end{eqnarray}
\begin{eqnarray}
[\erf(\frac{\mu_2 - \mu_1}{\sigma_B \sqrt{2}}) - \Theta(\Delta b +
\mu_2 - \mu_1) ] \; \times \; [\mu_1 - \mu_2 - \Delta b] . 
\label{eq:risk}
\end{eqnarray}
Say that $\Delta b = 0$. Then Eq.~\ref{eq:risk} shows that so long as
$\mu_1 \ne \mu_2$, as $\sigma_B \rightarrow 0$ risk goes to its
minimal possible value of zero. So everything else being equal,
shrinking the variance of either estimator reduces risk, essentially
minimizing it.  Alternatively, if we leave the variances of the two
estimators unchanged, but increase their covariance,
$\frac{1}{2}[\sigma^2_A - \sigma_B^2]$, then $\sigma_A$ will increase,
while $\sigma_B$ must shrink. So again, the risk will get reduced.
For the more general, non-Gaussian case, the high order moments may
also come into play.

\end{appendices}
\bibliography{pd,dbib,landscape.3,mathref}

\end{document}